\documentclass[10pt,journal,compsoc]{IEEEtran}
\usepackage{subfig}
\usepackage{graphicx}
\usepackage{multirow}
\usepackage[T1]{fontenc}
 \usepackage{hyperref}
\usepackage{color}

\usepackage{amssymb}
\usepackage{graphicx}
\usepackage{amsmath}

\newcommand{\real}{\ensuremath{\mathbb{R}}}
\newcommand{\ltwo}{\ensuremath{\mathbb{L}^2}}

\ifCLASSOPTIONcompsoc
  \usepackage[nocompress]{cite}
\else
  \usepackage{cite}
\fi
\ifCLASSINFOpdf
\else

\fi
\usepackage{amsmath}
\usepackage{algorithm, algpseudocode}
\newcommand{\etal}[1]{{\textit{et al.}{#1}}}
\newcommand{\ie}[1]{{\textit{i.e.}{#1}}}
\newcommand{\eg}[1]{{\textit{e.g.}{#1}}}
\definecolor{orange}{rgb}{0.6, 0.5, 0.0}
\newcommand{\SB}[1]{{\color{black}{#1}}}

\newcommand{\NO}[1]{{\color{black}{#1}}}

\begin{document}
%

\title{Dynamic Facial Expression Generation on Hilbert Hypersphere with Conditional Wasserstein Generative Adversarial Nets}

\author{Naima~Otberdout,~\IEEEmembership{Member,~IEEE,}
Mohamed~Daoudi,~\IEEEmembership{Senior,~IEEE,}
Anis~Kacem,~\IEEEmembership{Member,~IEEE,}
Lahoucine~Ballihi,~\IEEEmembership{Member,~IEEE,}
and~Stefano~Berretti,~\IEEEmembership{Senior,~IEEE}
\IEEEcompsocitemizethanks{\IEEEcompsocthanksitem N. Otberdout and L. Ballihi are with the LRIT - CNRST URAC 29, Mohammed V University in Rabat, Faculty of Sciences, Rabat, Morocco,
E-mail: naima.otberdout@um5s.net.ma, lahoucine.ballihi@um5.ac.ma 
\IEEEcompsocthanksitem A. Kacem and M. Daoudi are with IMT Lille-Douai, University of Lille, CNRS, UMR 9189 CRIStAL, Lille, France. E-mail:. \{anis.kacem,mohamed.daoudi\}@imt-lille-douai.fr
\IEEEcompsocthanksitem S. Berretti is with the Department of Information Engineering, University of Florence, 
Florence, Italy. E-mail: stefano.berretti@unifi.it}}

\IEEEtitleabstractindextext{%
\begin{abstract}
In this work, we propose a novel approach for generating videos of the six basic facial expressions given a neutral face image. 
We propose to exploit the face geometry by modeling the facial landmarks motion as curves encoded as points on a hypersphere. By proposing a conditional version of manifold-valued Wasserstein generative adversarial network (GAN) for motion generation on the hypersphere, we learn the distribution of facial expression dynamics of different classes, from which we synthesize new facial expression motions. The resulting motions can be transformed to sequences of landmarks and then to images sequences by editing the texture information using another conditional Generative Adversarial Network. To the best of our knowledge, this is the first work that explores manifold-valued representations with GAN to address the problem of dynamic facial expression generation. We evaluate our proposed approach both quantitatively and qualitatively on two public datasets; Oulu-CASIA and MUG Facial Expression. Our experimental results demonstrate the effectiveness of our approach in generating realistic videos with continuous motion, realistic appearance and identity preservation. We also show the efficiency of our framework for dynamic facial expressions generation, dynamic facial expression transfer and data augmentation for training improved emotion recognition models.


\end{abstract}

}

\maketitle

\IEEEdisplaynontitleabstractindextext

%
\IEEEpeerreviewmaketitle

\IEEEraisesectionheading{\section{Introduction}\label{sec:introduction}}
\begin{figure}[!ht]
\centering 
\includegraphics[width=\linewidth]{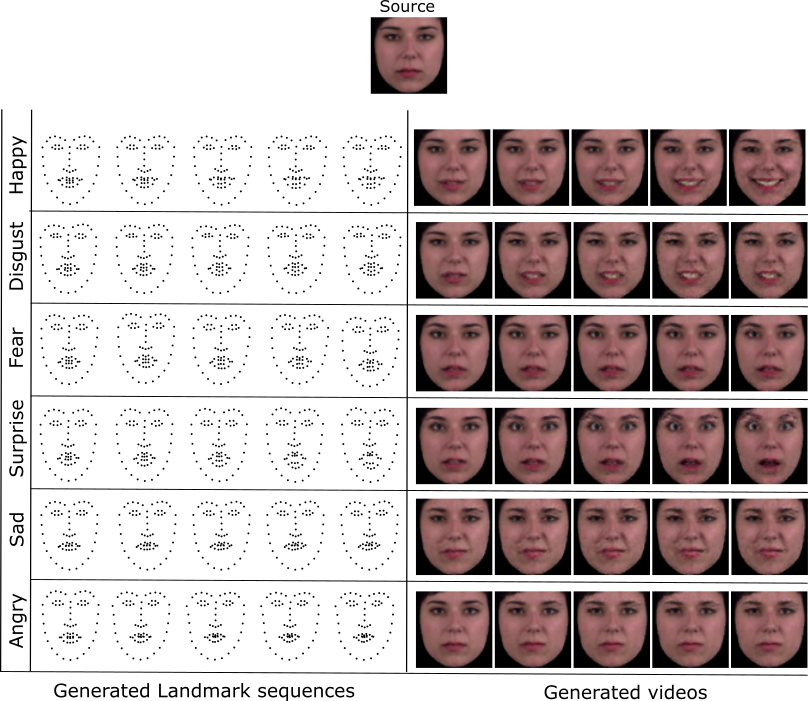}
\caption{Given a neutral image, our proposed model is able to generate sequences of facial landmarks for different facial expressions and transform them to videos.}
\label{pix2pixResults}
\end{figure}

\begin{figure*}[!ht]
\centering
\includegraphics[width=13.8cm]{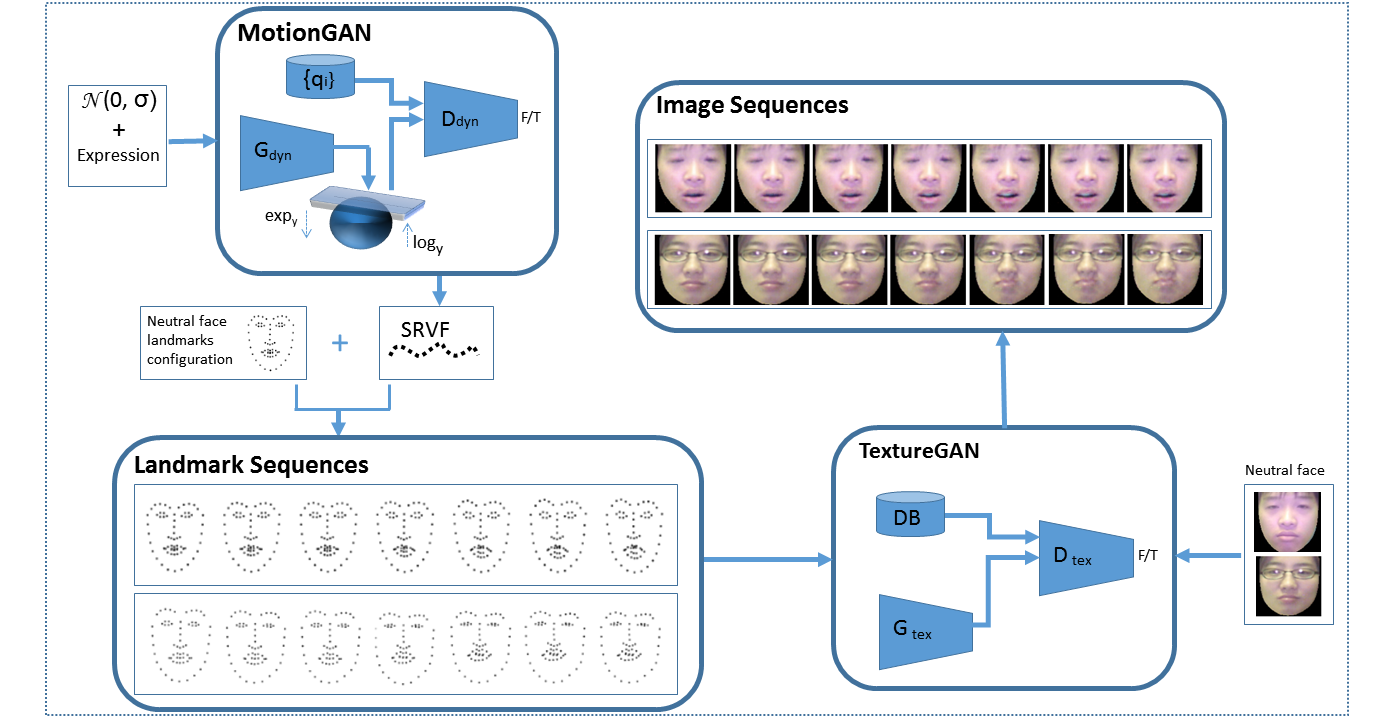}
\centering
\caption{Overview of the proposed approach. The proposed architecture consists of two GANs. First, MotionGAN synthesizes facial expression motion corresponding to the desired expression from noise. The resulting motion encoded by the Square-Root Velocity Function is then applied to neutral face landmark configurations to generate a sequence of landmarks. Finally, TextureGAN transforms the sequence of landmarks to a sequence of frames corresponding to the input identity.}
\label{Fig:Overview}
\end{figure*}

Since several decades, automated facial expression analysis has been studied in many computer vision researches~\cite{PanticIEEESMC2006, COHEN2003160}. Indeed, facial expressions play a vital role in social interaction and involve several potential applications that go from human computer interaction to medical and psychological investigations. For a long time, several works have tackled the problem of facial expression recognition, while the problem of facial expression generation is more challenging and has been less investigated in the state-of-the-art. 
\SB{Actually, synthesizing realistic facial expressions can have several applications, including, data augmentation to improve the performance of a facial expression classifier, facial expression transfer, face recognition (to train robust models for facial expressions), robot and avatars animation, computer games, facial surgery planning, etc.} 

Recently, with the success of Generative Adversarial Networks (GANs)~\cite{NIPSGoodfellow2014} for image generation, a big progress has been made in this task. These networks, that learn to produce samples similar to a given data distribution through a two-player game, have shown to be powerful in many image generation tasks including static facial expression synthesis~\cite{Ding2018}. However, the majority of the proposed solutions have addressed the problem of static facial expression generation, while the problem of dynamic facial expression synthesis is still less investigated. 
Given that facial expressions are much more described by a dynamic process than a static one, we need new solutions to synthesize realistic facial expression videos. This is a more challenging problem due to the difficulty of generating the dynamic evolution of facial expressions. 
Indeed, generating a facial expression video from a single photo is a one-to-many problem, where the output has much more unknowns to find than the input that does not include any temporal information. Besides, in this problem we not only need to generate realistic images, but we also have to generate continuous frames that change smoothly along time without sudden changes. 
The question that we addressed in this paper is, given one single neutral face, can we generate diverse face expression videos conditioned on a facial expression label as shown in Figure~\ref{pix2pixResults}? Responding to this question involves three tasks to be learned by the generative model: (1) the dynamic evolution of the facial expressions to synthesis video frames with continuous and smooth changes, which necessitates to define a robust and efficient motion representation, (2) the video appearance to synthesis realistic images, and (3) how to change the facial expression, while preserving the identity and its characteristics (\eg, eyeglasses, beard, etc.).

In this paper, we address these problems by proposing a new approach that utilizes a trajectory-based representation of landmark sequences on the hypersphere manifold to learn and generate video motions. The pipeline of the proposed framework is shown in Figure~\ref{Fig:Overview}. First, a compact representation of a landmarks sequence is encoded as points on the hyperphere manifold. Next, a new conditional version of Wasserstein GAN for manifold-valued data that we call \textbf{MotionGAN} is used to learn the distribution of the resulting representations. After training \textbf{MotionGAN}, we generate new facial expression motions corresponding to one of the six basic facial expressions. These motions can then be applied to any neutral facial landmark configuration to generate a sequence of facial landmarks. Finally, this sequence is fed to a Pix2Pix network~\cite{pix2pix2017}, and generates realistic textured facial expression videos. 

In summary, the novel aspects and contributions of this paper are as follows:
\begin{enumerate}
\item \textbf{Manifold-valued representations for GANs}: We represent a sequence of facial expression landmarks as curves that can be mapped to points on the hypersphere manifold. To the best of our knowledge, this is the first work that explores manifold-valued representations with GAN to address the problem of dynamic facial expression generation;
\item \textbf{A conditional manifold-valued Wasserstein GAN}: We generalize manifold-valued Wasserstein GAN to supervised generation by proposing a conditional version of this GAN; 
\item \textbf{Six dynamic facial expression synthesis from a single face}: By combining two disparate ideas GAN and Riemannian geometry tools, we propose a facial expression generation approach that achieves good results for dynamic facial expression editing, facial expression transfer and data augmentation.
\end{enumerate}

We evaluate our proposed approach both quantitatively and qualitatively on two public databases: the Oulu-CASIA and MUG Facial Expression datasets. Our experimental results demonstrate the effectiveness of our approach in generating realistic videos with continuous motion, realistic appearance and identity preservation. Besides, we show our approach can be used for dynamic facial expression transfer. We also demonstrate the usefulness of the generated data by using them for data augmentation to train a recurrent neural network classifier.
The rest of the paper is organized as follows: In Section~\ref{sect:related-work}, we review some recent works that have tackled the same or related problems. In Section~\ref{sec:problem}, we present and formulate our problem. Then, \textbf{MotionGAN} and \textbf{TextureGAN} details are provided in Section~\ref{sec:LandmarksGenerator} and~\ref{sect:TextureGAN}, respectively. In Section~\ref{sec:experiments}, we present an extensive experimentation of the proposed approach as well as its application to facial expression transfer and data augmentation; Lastly, conclusions and discussion are reported in Section~\ref{sect:conclusions}.

\section{Related work}\label{sect:related-work}
A big progress has been recently obtained in image generation using deep generative models, especially with GANs~\cite{NIPSGoodfellow2014}. Facial expressions editing is one of the numerous fields that have benefited from this progress. Indeed, many solutions have been proposed to synthesize facial expression images or to transfer facial expressions from one identity to another one using GANs. In this section, we review some of these solutions that we divided into two groups: Firstly, we review some recent advances achieved in static facial expression synthesis with GANs; Then, we discuss relevant works that tackled the problem of dynamic facial expression generation. Before discussing these works, we introduce GAN and its variants including the version generalized to manifold-valued data.

\textbf{Generative Adversarial Networks} -- Recently, GANs have shown to be extremely efficient for synthesizing realistic images. GANs have been used for several applications including image synthesis~\cite{radford2015unsupervised}, image-to-image translation~\cite{pix2pix2017, zhu2017unpaired, choi2018stargan}, image super-resolution~\cite{ledig2017photo, Chen:2018}, facial expression transfer~\cite{ding2018exprgan, song2018geometry, zhou2017photorealistic}, face aging~\cite{zhang2017age, yang2018learning}, face pose manipulation~\cite{yin2017towards}, etc.
The general idea of GANs consists of training two neural networks, generator and discriminator, in a minimax two-player game. The training process aims to learn the natural image distribution by forcing the generator to output samples that are indistinguishable from natural images. After the great success of GAN, several variants of its architecture have been proposed, including conditional GAN (CGAN)~\cite{mirza2014conditional}, that uses condition label to guide the generation process. Based on CGAN, Isola~\etal~\cite{pix2pix2017} proposed the pix2pix network that was widely used in the state-of-the-art~\cite{wang2018high, zhu2017unpaired}. It is a conditional GAN, where the generator is a U-Net~\cite{ronneberger2015u}, and the discriminator adopts a patch-based fully convolutional network~\cite{long2015fully}. This network uses an $L1$ reconstruction loss to enforce the generated samples to be locally close to the ground truth, and an adversarial loss to correct the blur effect of the $L1$ loss and synthesize realistic images. Theoretically, all the GAN architectures discussed above minimize Jensen-Shannon divergence between the true and the generated data distributions. By contrast, the Wasserstein GAN (WGAN)~\cite{pmlr-v70-arjovsky17a} minimizes a Wasserstein-1 distance between the two distributions \SB{making progress toward stable training of GANs. Other solutions further improved the training stability of standard WGAN by penalizing the norm of gradient of the critic with respect to its input~\cite{gulrajani2017improved}.}
While GAN techniques have shown a great success for real-valued image generation, they have been rarely applied to manifold-valued images. This idea has been discussed in~\cite{ZhiwuHuang2017}, where a generalization of WGANs for manifold-valued data in the context of static image generation was proposed. This poses several challenges such as the adaptation of cost functions and optimization algorithms to the manifold-valued data, and the generalization of the classic distribution distance to manifolds. To the best of our knowledge, our work is the first one that proposes a generalization of the Conditional WGAN for manifold-valued data to address the problem of dynamic facial expression generation.

\textbf{Facial Expression Image Generation} -- Conditional GANs~\cite{mirza2014conditional} have been exploited in many works to generate realistic face images corresponding to a given expression.
\SB{Some works pursued the general objective of generating a face with pose and expression that are under the control of a neural network model. For example, this approach was followed by Wiles \etal~\cite{wiles2018x2face} by using another face or a different modality (\eg, audio).}
Following a first research direction, some works integrated the target expression in a GAN as a deterministic one-hot vector that encodes the target class, thus generating faces conditioned on discrete emotion states~\cite{Zhang_2018_CVPR, lai2018emotion}. While this solution can be considered more simple, it generates only discrete facial expressions, which significantly reduces the diversity of the generated samples. To tackle this issue, Ding \etal~\cite{ding2018exprgan}, proposed an Expressive GAN (ExprGAN) for facial expression editing that uses an expression controller module. This module is a real-valued vector conditioned on the label that encodes more complex information about the target expression such as its intensity variation.
\SB{Pumarola \etal~\cite{pumarola2018ganimation} introduced a GAN conditioning scheme based on Action Units (AU) annotations, which describes in a continuous way the anatomical facial movements defining a human expression. This made possible to control the magnitude of activation of each AU and combine several of them.}
\SB{Some methods used 3D priors to animate static images. 
For example, Kollias \etal~\cite{kollias2020deep} fit a 3D Morphable Model (3DMM) to a target neutral image. In this solution an input affect is added to the 3DMM and the new face is blended with the given affect into the target image. The approach allows synthesizing facial affect either in terms of the six basic expressions, or in terms of how positive or negative is an emotion (\textit{valence}), and power of the emotion activation (\textit{arousal}). 
Ververas and Zafeiriou~\cite{ververas2019slidergan} proposed SliderGAN for image-to-image translation, under a set of continuous parameters. This allows transforming an input face image into a new one according to the continuous values of a statistical blendshape model of facial motion. Facial image editing is shown for expression and speech blendshapes.}

Following a different direction, other works take advantage of facial geometry and encode the target expression through facial landmark positions~\cite{song2018geometry, Qiao:2018}. This solution is more flexible, allowing the synthesis of continuous facial expressions, and can be exploited for facial expression transfer as well. The main idea consists of encoding facial landmark positions in heatmaps that provide a per-pixel likelihood for these point locations. These heatmaps are usually fed to the GAN as an additional channel concatenated with the face image to synthesize this face performing the expression encoded in the heatmap. This strategy was adopted by many works including Song \etal~\cite{song2018geometry} who proposed a geometry-guided GAN architecture for facial expression editing and removal. The face geometry is used as a condition introduced in the network to control the process of expression generation or expression removal. In the same direction, Qiao \etal~\cite{Qiao:2018} proposed a Geometry-Contrastive GAN (GC-GAN) to transfer facial expressions across different identities using face geometry. 
\SB{Songsri-in and Zafeiriou~\cite{songsri2019face} proposed an image-to-image translation that takes a facial image, its current landmarks, and target landmarks as inputs and generates a full facial image sequence encoded by the target landmarks.}
Motivated by the ideas discussed above, we also take advantage of the face geometry by encoding the expression in a landmarks configuration. However, unlike previous works that generate static images of facial expressions, we tackle the problem of facial expression video synthesis, which is more challenging because of the temporal dynamics of the video that has to be captured and generated.

\textbf{Video Generation with GAN} -- Despite the remarkable success achieved in image synthesis, video generation is still a more challenging task, due to the difficulty of generating the temporal motion of the video. Some recent works have started to explore GANs for video generation. 
\SB{For example, Wang \etal~\cite{wang2018video} proposed a video-to-video synthesis approach under the generative adversarial learning framework. High-resolution, photorealistic, temporally coherent videos were generated on a set of input formats including segmentation masks, sketches, and poses. 
Kim \etal~\cite{kim2018deep} proposed a generative neural network for photo-realistic re-animation of portrait videos using only an input video. In doing so, they transfer the full 3D head position, head rotation, face expression, eye gaze, and eye blinking from a source actor to a portrait video of a target actor.}
Other works tackled the problem of future frame prediction~\cite{fan2018controllable, zhou2016learning, oh2015action}, with some techniques capable of generating a whole video starting from one image. 
These latter works can be roughly divided into two groups according to the way they handle and generate the video motion.
In the first group, a spatio-temporal network is used to generate all video frames at the same time. This category includes the methods in~\cite{NIPS2016_6194} and~\cite{MasakiSaito2017} that exploit 3D spatio-temporal GANs to generate multiple frames of the video simultaneously.
\SB{Using only a still image of a person and an audio clip containing speech, in~\cite{vougioukas2019realistic} Vougioukas \etal~proposed to generate videos of a talking head with lip movements that are in sync with the audio. Generated videos also show blinks and eyebrow movements. This has been obtained using a temporal GAN with three discriminators focused on achieving detailed frames, audio-visual synchronization and realistic expressions.}
However, these approaches usually synthesize videos with poor image quality. To tackle this issue, other works explored different directions and proposed to generate the video motion separately, then generate the video frames sequentially. This is achieved by using Recurrent Neural Networks (RNNs). In~\cite{Tulyakov2018}, Tulyakov \etal~proposed the MoCoGAN framework for video generation. The idea consists of decomposing the video into content and motion information, where the video motion is learned by Gated RNN (GRU) and the video frames are generated sequentially by a GAN. This approach was experimented in different applications including dynamic facial expression synthesis. However, the generated images presented content and motion artifacts and failed to capture fine details of facial expressions.  \NO{Indeed, generating the motion as a sequence of separate feature vectors with GRNN, results in motion artifacts and discrete transitions. By contrast, we propose in this paper to generate a compact motion as one point on the hypersphere, which minimizes these motion artifacts and results in smoother motions.}
In~\cite{WeiWang2018}, Wang \etal exploited facial landmarks for generating smile videos. They also modeled the temporal evolution of the smile expression using an RNN that produced landmark sequences translated later to image sequences using GAN. This work generates videos with acceptable image quality, but they focused only on smile expression that is more simple to capture and generate compared to other facial expressions.
Following these last approaches, we also guide the generation process of dynamic facial expression by facial landmark sequences. However, different from the previous works, instead of directly generating facial landmark sequences, we generate the motion of these landmarks that can then be applied to any facial landmarks configuration to generate videos of different identities performing this motion. This is achieved by modeling landmark sequences as curves that can be mapped to points on a manifold space. We explicitly leverage the geometry of this manifold to generate the six facial expression sequences as new points on the manifold. To the best of our knowledge, this is the first study exploring manifold-valued representations to generate video motions with GAN, while common solutions to this task in the state-of-the-art were based on RNN.

\section{Problem Statement}\label{sec:problem}
Given a neutral face image $X_0$ with its landmarks $Z_0$, and a desired expression $c$, we aim to generate a facial expression sequence $\{X_1,...,X_T \}$ of length $T$ corresponding to the identity of $X_0$ and the expression $c$. To achieve this goal, we seek a mapping $\eta$ between the input $\{X_0, Z_0, c\}$ and its associated facial expression sequence $\{X_1,...,X_T \}$ . To simplify the complexity of the task at hand, we define this mapping as a composition of two functions $\eta = \theta \circ \phi$, such that the function $\phi$ learns the distribution of facial expression dynamics to generate the temporal evolution of the sequence, while $\theta$ learns how to synthesize its texture information. Since the visual signal in a video can be divided into content (texture) and dynamics, such decomposition facilitates the problem of video generation and allows us to design functions that focus on learning less complicated tasks. Moreover, this decomposition allows applying the same generated dynamics to different identities to generate videos of different persons performing the same facial expression. It is also possible to apply different dynamics to the same identity to generate videos of the same person performing different facial expressions. 

In order to define the function $\phi$, we completely ignore the texture information and focus only on facial landmarks. Let us consider a set of $m$ training image sequences $\mathcal{X}=\{\{X_1^j,...,X_T^j \}\}_{j=1}^m$ and its corresponding set of $m$ sequences of landmark configurations $\mathcal{Z}=\{\{Z_1^j,...,Z_T^j \}\}_{j=1}^m$, such that $Z_i, 1\leq i\leq T$ is the landmarks configuration of the facial image $X_i, 1\leq i\leq T$. From the set $\mathcal{Z}$ of landmark configuration sequences, the function $\phi$ can learn the distribution of facial expressions dynamics. The key idea here is to model the temporal evolution of landmark configurations as time-dependent 2D curves, which can be efficiently represented as single and compact points on the hypersphere manifold~\cite{SrivastavaPami2011}. By doing so, the landmark sequences are considered as manifold-valued data, and the geometric properties of this manifold are exploited to define a manifold-valued GAN \textit{MotionGAN} for learning the distribution of facial expressions dynamics $\phi$. As manifold-valued data lie on Riemannian manifolds rather than Euclidean space, the definition of manifold-valued data distribution is different from that of real-valued data distribution; then, it is unfeasible to apply traditional GANs directly in this case. This is due to the fact that traditional GANs generate new samples that do not lie on the hypersphere manifold in contrast to the training real data dynamics that are represented on the latter manifold. By proposing a conditional version of the manifold-valued Wasserstein GAN introduced in~\cite{ZhiwuHuang2017}, which is defined on the hypersphere manifold, our approach learns the manifold-valued distribution $\phi$ and generates the dynamics of new facial expressions that can be used to generate a new sequence of landmarks $\{Z_1^{new}, \dots , Z_T^{new}\}$.  
Finally, another GAN, called \textit{TextureGAN} is used to define the function $\theta$ that learns from training data $\mathcal{X}$ and $\mathcal{Z}$ to translate the sequence of landmark configurations $\{ Z_1^{new}, \dots , Z_T^{new}\}$ to a sequence of video frames $\{ X_1^{new}, \dots , X_T^{new}\}$. 

The overall architecture illustrated in Figure~\ref{Fig:Overview} consists of three blocks. First, MotionGAN learns the motion of landmarks from facial landmark sequences of the training set and generates new conditional facial expressions motions. The second block generates a sequence of landmarks corresponding to the facial expression dynamics generated by MotionGAN and a neutral landmark configuration. The last block translates the landmark sequence to a video: it receives a neutral face image and a sequence of landmarks and produces their corresponding facial expression video. 
In the following sections, we will provide more details about each one of these blocks.

\section{MotionGAN: A Conditional manifold-valued GAN for Facial Expression motions Generation}\label{sec:LandmarksGenerator}
Facial landmarks have been widely used for facial expression analysis~\cite{kacem2018novel, jung2015joint}. Indeed, facial landmarks are considered as a powerful tool to capture the geometric features of the face and encode both the appearance and the dynamic of the facial expression. Besides, facial landmarks have been recently adopted as guiding information in facial expression synthesis in several works~\cite{song2018geometry, WeiWang2018}. The idea here consists of using facial landmarks as an additional image channel to be concatenated with the input face directly or as a vector of landmark coordinates to be used as a guide during the generation process. 

Based on this motivation, we also take advantage of the geometric information provided by facial landmarks in two ways. On the one hand, we exploit the evolution of landmark locations to encode the dynamics of the face. This is achieved by modeling the landmarks evolution as a curve that can be represented and analyzed in a Riemannian manifold. By exploiting these curves, we train a GAN to generate new facial dynamics that we can transform to new dynamic landmark configurations of any face given its neutral landmarks configuration. On the other hand, following the state-of-the-art, we use the generated landmark configurations to guide the generation of the final video frames by adding the texture information in the input image of the GAN to the geometric features provided by landmarks. 

In this section, we present more details about the first stage of our method, which consists of modeling, analyzing and generating dynamic facial landmarks using MotionGAN. We first introduce the geometric framework used for data modeling, analysis and representation in a Riemannian manifold. Then, we present the architecture and the details about the learning process of MotionGAN.

\subsection{MotionGAN Data Modeling}
Let us represent a facial expression video of $T$ frames by its corresponding sequence of $T$ landmark configurations $Z=\{Z_1,\ldots, Z_{T}\}$, where each configuration $Z_{i}, 1 \leq i\leq T$ is a $2 \times d$ matrix of rank $d$ encoding the 2D positions of $d$ distinct landmark points $\{p_i=(x_i, y_i)\}_{i=1}^d$. Following such representation, we model the sequence $Z$ as a curve represented by a continuous parameterized function $\beta(t): [0,1] \rightarrow \mathbb{R}^{2d\times T}$. These representations allow us to widely simplify the problem of landmark sequences generation given that each curve can be mapped to one single point on a given manifold. 
More formally, each curve $\beta(t)$ can be represented by its \textit{Square-Root Velocity Function} (SRVF)~\cite{SrivastavaPami2011}, $q : [0,1] \rightarrow \mathbb{R}^{2d\times T}$ according to,
\begin{equation}
q(t) = \frac{\dot{\beta}(t)}{\sqrt{ \|\dot{\beta}(t)\|}} \; ,
\label{curve_2_q}
\end{equation}

\noindent
where $\|\cdot\|$ is the $\mathbb{L}^2$-norm in $\mathbb{R}^{2d\times T}$. The effectiveness of such specific representation for shape analysis has been proven in 3D facial curves~\cite{DriraASDS13} and action recognition~\cite{Devanne2015}. This representation encodes the temporal evolution of the facial landmark configurations and so the dynamics of the facial expression.  In~\cite{SrivastavaPami2011}, authors proposed to remove the scale variability of the resulting curves by scaling the $\mathbb{L}^2$-norm of these functions to $1$ (\ie, $\|q\|=1$). Accordingly, the space of the resulting representations $\mathbb{S}=\{q:[0,1] \rightarrow \mathbb{R}^{2d\times T}, \|q\|=1\}$ becomes a unit hypersphere in the Hilbert manifold $\ltwo([0,1],\real^{2d\times T})$, and each landmark configuration sequence becomes a point on this spherical manifold. Consequently, we reduce the problem of landmark sequences generation to a problem of generating points on the spherical manifold $\mathbb{S}$. 

The first step needed in analyzing and comparing these representations consists of parametrizing them. Indeed, due to the different execution rates of the facial expressions, aligning and parametrizing these curves is a crucial processing for efficiently compare them. Formally, given two curves $\beta_1$ and $\beta_2$, their corresponding SRVFs $q_1$ and $q_2$ can be registered by finding the non-linear function $\gamma: [0,1] \rightarrow [0,1]$, $\gamma^{*} = \underset{{\gamma \in \Gamma}}{\operatorname{argmin}}(\| q_1 - \sqrt{\dot{\gamma}} q_2 \circ \gamma \|)$, which optimally registers the two curves allowing their rate-invariant comparison. The optimal parametrization function $\gamma^{*}$ can be found by using a Dynamic Programming algorithm as explained in~\cite{SrivastavaPami2011}. After registration, we can compute efficiently the distance $d_\mathbb{S}$ between the two registered curves $q_1$ and $q_2^*=q_2\circ \gamma^*$ according to, 
\begin{equation}
\label{eq:distance_sphere}
d_{\mathbb{S}}(q_1,q_2^*)=\cos^{-1}(\langle q_1,q_2^* \rangle) \; . 
\end{equation}

\noindent
This distance quantifies the similarity between the two curves in $\mathbb{R}^{2d\times T}$. As explained in~\cite{SrivastavaPami2011}, this distance is invariant to rotation and scaling, and it also considers the stretching and the bending of the curves. In our approach, we also need to define the statistical mean of these representations in order to define a representative element of a specific group (\eg, a representative curve of the happy expression). To this end, we introduce the Riemannian center of mass, also known as Karcher mean~\cite{karcher77}, which can be used to compute an average element of a set of points in the hypersphere manifold. More formally, we define the Karcher mean $q_{mean}$ of a set of points $\{q_i\}_{i=1}^N$ in the hypersphere manifold $\mathbb{S}$ according to $q_{mean}=\underset{{q_i} \in \mathbb{S}}{\operatorname{argmin}} \sum_{i=1}^N d_{\mathbb{S}}(q, q_i)^2$. 
In our work, we compute the Karcher mean to derive a representative curve of each facial expression, which is used to align the other curves. We also exploit the Karcher mean to define a reference point $y$, where we define the tangent space $T_y(\mathbb{S})$ of $\mathbb{S}$ that will be used in the training of \textit{MotionGAN}. In the next section, we show how to use the mathematical tools introduced here to train \textit{MotionGAN} to generate new points in $\mathbb{S}$ that encode the dynamics of new facial expressions.

After defining the mathematical tools used to analyze and generate facial expression dynamics in the hypersphere manifold $\mathbb{S}$, we need to recover the landmarks configuration sequence $Z_{new}$ that corresponds to a new generated point $q_{new}$ in $\mathbb{S}$. Conveniently to us, for each $q \in \mathbb{S}$ there exists a unique curve $\beta$ up to a translation such that the given $q$ is the SRVF of that $\beta$. Formally, the curve $\beta$ can be recovered within a translation, using,
\begin{equation}
\beta(t) = \int_{0}^{T}\|q(s)\| q(s) ds + \beta(0) \; ,
\label{eq:q_2_curve}
\end{equation}

\noindent
where $\beta(0)$ represents the landmarks configuration $Z_0$ of the initial frame. According to this equation, we can apply the generated facial expression dynamics encoded in $q$ to any identity. Indeed, by using the landmarks configuration of any identity as an initial condition in~\eqref{eq:q_2_curve}, we can recover the sequence of landmark configurations corresponding to this identity performing the motion encoded in $q$.  

\subsection{MotionGAN Network}
Given a set of $m$ training samples of facial landmark configuration sequences $\mathcal{Z}=\{(\{Z_1^j,,...,Z_T^j \}, c^j)\}_{j=1}^m$ with their associated facial expression classes, we compute their corresponding parametrized SRVFs set $\mathcal{Q}=\{(q^j,c^j)\}_{j=1}^m$ to train the \textit{MotionGAN} model.

Given that SRVF representations are manifold-valued data that lie on Riemannian manifolds rather than Euclidean space, we propose MotionGAN, a conditional version of Wasserstein GAN for manifold-valued data to learn the distribution of SRVFs associated to each emotion class. This GAN is an extended version of CGAN from the Euclidean space to the hypersphere manifold $\mathbb{S}$. We exploit the logarithm and exponential maps defined for the hypersphere manifold $\mathbb{S}$, given later by~\eqref{eq:LogSphere} and~\eqref{eq:ExpSphere}, respectively, to optimize the Wasserstein distance between the distribution of the generated manifold-valued data and that of the real manifold-valued data under an adversarial training.
MotionGAN maps a random vector $\mathbf{z}$ to an SRVF point on $\mathbb{S}$. 
It consists of two adversarial models: a generative model $G_{dyn}$ that captures the data distribution $\mathbb{P}_{dyn}$ of the expressions dynamics encoded in the SRVFs $q$, and a discriminative model $D_{dyn}$ that estimates the probability that a sample come from the training data rather than the generator $G_{dyn}$. The goal of training these two models is to learn a function $\phi : \mathbb{R}^{n} \rightarrow \mathbb{S}$, which maps an $n$-dimensional noise vector sampled from a normal distribution $z \sim \mathbb{P}_z$ to an SRVF $q \in \mathbb{S}$ encoding the dynamic evolution of landmarks. 

\begin{figure*}
\begin{center}
\includegraphics[width=12.33cm]{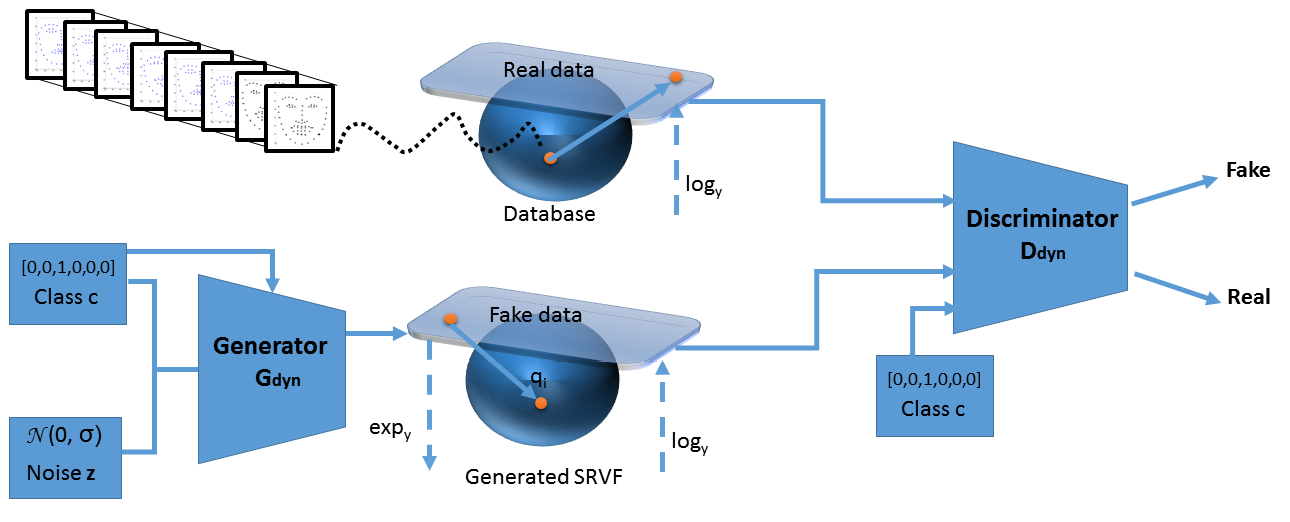}
\end{center}
\caption{Overview of MotionGAN, the Conditional Wasserstein GAN used for motion generation.}
\label{Fig:ShapeGANArchi}
\end{figure*}

\subsection{Loss Function}
The global objective function used to train MotionGAN is a weighted sum of three loss functions: the adversarial loss $L_{adv}$, the reconstruction loss $L_{S\_recon}$ in $\mathbb{S}$, and the reconstruction loss $L_{T\_recon}$ in $\mathcal{T}_y(\mathbb{S})$ such that,
\begin{equation}
\label{eq:MotionGAN_loss}  
L_{MotionGAN}= \alpha_1 L_{adv} + \alpha_2 L_{S\_recon} + \alpha_3 L_{T\_recon} \; .
\end{equation}

\noindent
Regarding the adversarial loss, we propose the conditional version of the objective function proposed in~\cite{ZhiwuHuang2017}. This function is the generalized version of the objective function of Wasserstein GAN~\cite{pmlr-v70-arjovsky17a} to the hypersphere manifold-valued data, according to,
\begin{equation}
\begin{array}{rl}
L_{adv}= & {\mathbb{E}_{\boldsymbol{q} \sim \mathbb{P}_{dyn}}\left[D_{dyn}\left(\log _{\boldsymbol{y}}(\boldsymbol{q}),c\right)\right]} 
\\ 
{ } & {-\mathbb{E}_{G_{dyn}(\boldsymbol{z}) \sim \mathbb{P}_{g}}\left[D_{dyn}\left(\log _{\boldsymbol{y}}\left(\exp _{\boldsymbol{y}}(G_{dyn}(\boldsymbol{z}, c))\right)\right)\right]}
\\ 
{ } & {+\lambda \mathbb{E}_{\hat{\boldsymbol{q}} \sim \mathbb{P}_{\hat{\boldsymbol{q}}}}\left[\left(\left\|\nabla_{\hat{\boldsymbol{q}}} D_{dyn}(\hat{\boldsymbol{q}})\right\|_{2}-1\right)^{2}\right]},
\end{array}
\label{Eq:LossWassersteinGAN}
\end{equation}

\noindent 
where $\log_y(.)$ and $\exp_y(.)$ are the logarithm and exponential maps, respectively, defined for the hypersphere manifold in a particular point $y$. The logarithm map $\log_{\boldsymbol{y}}(.)$ projects the SRVF $q$ from the hypersphere $\mathbb{S}$ to its tangent space $T_y(\mathbb{S})$ in $y$. This tangent space is a vector space, where any deep network can be applied directly, while the exponential map $\exp_{\boldsymbol{y}}(.)$ transforms the data back to the hypersphere manifold $\mathbb{S}$. The logarithm $\log_{\boldsymbol{y}}(.)$ and exponential $\exp_{\boldsymbol{y}}(.)$ maps for the hypersphere manifold $\mathbb{S}$ are defined by:
\begin{equation}
\label{eq:LogSphere}
\log _{\boldsymbol{y}}(q) = \frac{d_{\mathbb{S}}(q,y)}{sin(d_{\mathbb{S}}(q,y))} (q - cos(d_{\mathbb{S}}(q,y))y) \; ,
\end{equation}
\begin{equation}
\label{eq:ExpSphere}
\exp _{\boldsymbol{y}}(v)=cos(\|v\|)\boldsymbol{y} + sin(\|v\|)\frac{v}{\|v\|}.
\end{equation}

\noindent 
where $d_{\mathbb{S}}(q,y)$ represents the distance between $q$ and $y$ in $\mathbb{S}$ defined by~\eqref{eq:distance_sphere}.

In~\eqref{Eq:LossWassersteinGAN}, $\boldsymbol{z}$ is a random noise, and $\hat{\boldsymbol{q}}$ is a random sample following the distribution $\mathbb{P}_{\hat{\boldsymbol{q}}}$, which is sampled uniformly along straight lines between pairs of points sampled from the real distribution $\mathbb{P}_{dyn}$ and the generated distribution $\mathbb{P}_{g}$. It is given by,
\begin{equation}
\hat{\boldsymbol{q}} = (1 - \tau) \log_y(q) + \tau\log_y(\exp_y(G_{dyn}(z, c))), 
\end{equation}

\noindent
and $\nabla_{\hat{\boldsymbol{q}}} D_{dyn}(\hat{\boldsymbol{q}})$ is the gradient with respect to $\hat{\boldsymbol{q}}$. 
\NO{The last term in~\eqref{Eq:LossWassersteinGAN} represents the gradient penalty, which is added to the basic Wasserstein GAN. This was proposed in~\cite{gulrajani2017improved} to enforce the Lipschitz constraint and help to stabilize the training of the WGAN by penalizing the norm of the gradient of the discriminator.}

In addition to the adversarial loss, we use the two reconstruction losses $L_{T\_recon}$ and $L_{S\_recon}$. $L_{T\_recon}$ measures the distance in the tangent space $T_y(\mathbb{S})$ between the tangent vector $\log_y(q_{gt})$ of the ground truth SRVF $q_{gt}$ in $T_y(\mathbb{S})$ and its associated reconstructed vector $\log_y(\exp_y(G_{dyn}(z, c)))$. While $L_{S\_recon}$ quantifies the similarities between the generated SRVF $\exp_y(G_{dyn}(z, c))$ and its corresponding ground truth $q_{gt}$. The reconstruction loss in the tangent space is given by,
\begin{equation}
\label{eq:T_reconstruction_loss}
L_{T\_recon} = \|{ \log_{y}(\exp_y({G_{dyn}(z,c)})) - \log_y(q_{gt})} \|_1 \; ,
\end{equation}

\noindent
where $\|.\|_1$, represents the $L_1$-norm. The reconstruction loss on the hypersphere is given by,
\begin{equation}
\label{eq:S_reconstruction_loss}
L_{S\_recon}=d_{\mathbb{S}}( \exp_y({G_{dyn}(z,c)}) - q_{gt}) \; ,
\end{equation}

\noindent
where $d_{\mathbb{S}}$ is the geodesic distance in $\mathbb{S}$ given by~\eqref{eq:distance_sphere}. 

According to the objective function, the optimization of \textit{MotionGAN} is done in the tangent space $T_y(\mathbb{S})$ of the hypersphere $\mathbb{S}$ in a reference point $y$. Since the tangent space is a linear vector space, any regular network can be directly applied in $T_y(\mathbb{S})$. This is achieved by exploiting the logarithm map that maps the SRVFs $q$ of the database from $\mathbb{S}$ to $T_y(\mathbb{S})$, which forms the real data $\log(q)$ of the network. On the other hand, the exponential map transforms the generated fake data $G_{dyn}(z,c)$ from $T_y(\mathbb{S})$ to $\mathbb{S}$, then the logarithm map transforms the data back to the tangent space to force the generator to generate data on the desired tangent space of the sphere, which is $T_y(\mathbb{S})$. The discriminator $D_{dyn}$ takes as input the real data $\log_y(q)$ and the generated data $\log_y(\exp_y(G_{dyn}(z,c)))$, both laying on the tangent space of the sphere, and tries to distinguish between them. At the end of the training, the generator learns the distribution $\mathbb{P}_{dyn}$ of the real data and generates data similar to the real one on the desired tangent space of the sphere. 
The tangent space used in our training corresponds to the tangent space of the hypersphere in the Karcher mean of the data. In the following, we use the tangent space of the hypersphere to refer to the tangent space of the hypersphere computed in the Karcher mean of the training data.

After training, we use the resulting $G_{dyn}$ to generate a point on the tangent hypersphere conditioned on the desired expression. Then, we use the exponential map to find its corresponding point on the sphere, which represents an SRVF encoding a dynamic evolution of the desired expression. Finally, we use~\eqref{eq:q_2_curve} with a neutral landmarks configuration of any identity to generate the sequence of landmark configurations corresponding to this identity performing the desired expression encoded in the generated SRVF. In the following, we present the module that transforms the sequence of landmarks to a sequence of frames. 
 
Algorithm~\ref{algo:MotionGAN_training} outlines the steps used to train \textit{MotionGAN}, while Algorithm~\ref{Algo:landmarks_generation} summarizes the steps needed to generate a new sequence of facial landmark configurations using the trained \textit{MotionGAN}.

\begin{algorithm}[!ht]
\caption{Shape Conditional Wasserstein GAN training}
\label{algo:MotionGAN_training}
\hspace*{\algorithmicindent}
\textbf{Input}: $\{q_i, c_i\}_{i=1}^{m}$, training data with their corresponding labels; $w_0$, initial discriminator parameters; $u_0$, initial generator parameters; $a$, learning rate; $M$, batch size; $n_{disc}$, discriminator iterations per generation iteration; $\lambda$, balance parameter of gradient norm penalty; $n_{iteration}$, number of iterations for the generator. 
\begin{algorithmic}[1]
    \For{$i=1 \dots n_{iteration}$ }
        \For{$j=1 \dots n_{disc}$ }
            \State Sample minibatch of $M$ noise samples $\{z (1), \dots , z(M)\}$ from noise prior $\mathbb{P}_{z}$
            \State Sample minibatch of $M$ examples $\{q (1), \dots , q(M)\}$ from real data distribution $\mathbb{P}_{dyn}$
            \State Compute $\Delta_w$ the stochastic gradient of Eq.~\eqref{Eq:LossWassersteinGAN} with respect to $w$
            \State $w \leftarrow w + a$ . \textbf{AdamOptimizer}$(w, \Delta_w)$
        \EndFor
        \State Sample minibatch of $M$ noise samples $\{z (1), \dots , z(M)\}$ from noise prior $\mathbb{P}_{z}$
        \State Compute $\Delta_{u}$ the stochastic gradient of, $${- D_{dyn}^w\left(\log _{\boldsymbol{y}}\left(\exp _{\boldsymbol{y}}(G_{dyn}^{u}(\boldsymbol{z}, c))\right)\right)},$$ with respect to $u$
        \State $u \leftarrow u + a . \textbf{AdamOptimizer}(u, \Delta_{u}) $
    \EndFor
\end{algorithmic}
\end{algorithm}

\begin{algorithm}[!ht]
\caption{Landmarks Sequence Generation}
\label{Algo:landmarks_generation}
\hspace*{\algorithmicindent}
\textbf{Input}: $G_{dyn}$, Generator trained with Algorithm~\ref{algo:MotionGAN_training}; $Z_0$, neutral landmarks configuration; $c$ desired facial expression. 
\begin{algorithmic}[1]
    \State Sample $z$, a random noise from noise prior $\mathbb{P}_{z}$
    \State Generate $p=G_{dyn}(z, c)$, a point on the tangent space of the hypersphere $\mathbb{S}$
    \State Generate $q=\exp_y(p)$ by mapping the generated point $p$ to the hypersphere $\mathbb{S}$ using Eq.~\eqref{eq:ExpSphere}
    \State Generate sequence of landmarks using Eq.~\eqref{eq:q_2_curve} and $Z_0$ as initial condition
\end{algorithmic}
\end{algorithm}

\section{TextureGAN: Landmarks Sequence to Video Generation}\label{sect:TextureGAN}
After learning the distribution of expressions dynamics from the landmarks, we need to add the texture information to synthesize the final expression video. To this end, we use $TextureGAN$, which is a conditional GAN for real-valued data. This GAN uses a map of landmarks configuration $Z$ as a guide to generate the face of the input image with the expression corresponding to $Z$. Since the temporal information has been already tackled with MotionGAN, TextureGAN focuses only on the texture information in individual static images. However, even if it generates static images, its role is more complicated than generating a face with a certain expression as done in other static state-of-the-art approaches~\cite{ding2018exprgan} that focus on generating six or seven basic expressions. Indeed, TextureGAN has to generate also the intermediate frames of the video, that do not necessarily correspond to any of the basic expressions. Moreover, the generated frames have to show smooth continuous changes along time without sudden transitions. To this end, we guide the generation of expressions using landmarks: this simplifies the synthesis of intermediate frames and allows us to derive continuous expressions with smooth changes along time. This result is not possible, for example, when using a condition in the form of one-hot vector. 

Given a set of $m$ training image sequences $\mathcal{X}=\{\{X_1^j, \dots, X_T^j \}\}_{j=1}^m$ and its corresponding set of $m$ sequences of landmark configurations $\mathcal{Z}=\{\{Z_1^j, \dots, Z_T^j \}\}_{j=1}^m$, such that $Z_i, 1\leq i\leq T$ is the landmarks configuration of the facial image $X_i, 1\leq i\leq T$. Let $X_0^j$ denote the input neutral face image corresponding to the $j$-th training sequence. These sets will be used to train the generator $G_{tex}$ and the discriminator $D_{tex}$ of TextureGAN to learn the mapping $\theta$ between $X_0^j$ and $\{X_1^j, \dots, X_T^j \}$. To this end, we exploit a combination of a reconstruction and an adversarial losses. The global objective function of TextureGAN is a weighted sum of three loss functions; adversarial loss $L_{ADV}$, identity loss $L_{ID}$ and reconstruction loss $L_{REC}$ such that,
\begin{equation}
L_{TextureGAN}=\zeta_{1}L_{ADV} + \zeta_{2}L_{REC} + \zeta_{3}L_{ID}  \; ,
\end{equation}

\noindent
where the adversarial loss is given by,
%
\begin{equation}
\begin{array}{rl}
L_{ADV} = &\sum_{j=1}^{m} \sum_{t=1}^{T}\left[\log(D_{tex}(X_t^j,Z_t^j))\right] +
\\ 
{} & \sum_{j=1}^{m} \sum_{t=1}^{T}\left[\log(1- D_{tex}(G_{tex}(X_0^j),Z_t^j))\right]
\end{array}
\end{equation}

To further keep the face identity in the generated frames, we make use of an identity loss that enforces the similarity of identity features between the input and the output faces. To this end, we exploit the VGG-face~\cite{parkhi2015deep} model trained for face recognition to extract identity features, and maximize similarities between them. The identity loss is given by:
\begin{equation}
L_{ID} = \sum_{j=1}^{m}\sum_{t=1}^{T}\sum_{i=1}^5  \|{F_i(G_{tex}(X_0^j, Z_t^j)),F_i(X_t^j)   }\|_1 \; ,
\end{equation}

\noindent
where $\|.\|_1$ represents the L1 norm and $F_i$ are the features extracted from the $i$-th convolutional layer of the VGG-face. The used layers are conv1, conv2, conv3, conv4 and conv5. In order to keep the generated frames close to the ground-truth, we add a reconstruction loss to the global objective function. The reconstruction loss is given by:
\begin{equation}
\label{eq:T_reconstruction_loss2}
L_{REC} = \sum_{j}^{m}\sum_{t=1}^{T}\| X_t^j - G_{tex}(X_0^j,Z_t^j) \|_1 \; .
\end{equation}

\section{Experiments}\label{sec:experiments}
We performed several experiments to evaluate our approach. We first describe the used benchmarks and the experimental setup. Then, we present a quantitative and a qualitative evaluation for each part of our method including motion generation and video synthesis. We also introduce an ablation study to show the importance of each component of our approach. Finally, we evaluate our proposed method in two applications: data augmentation for expression recognition and dynamic expression transfer.

\subsection{Datasets} 
\textbf{Oulu-CASIA}~\cite{zhao2011facial}: This dataset contains over 480 videos of 80 subjects. For Each subject there there are six videos corresponding to the basic emotion labels; All videos begin with a neutral expression and end with the apex of the corresponding expression. The fist $80\%$ of the subjects was used for training, while the last $20\%$ was used as test set.

\noindent  \textbf{MUG Facial Expression}~\cite{aifanti2010mug}: This database includes videos of 86 subjects. Each video consists of 50 to 160 frames. We used only the sequences representing one of the six basic facial expressions, \emph{i.e.}, anger, disgust, fear, happiness, sadness, and surprise. Following~\cite{zhao2018learning}, we split the dataset into three parts in a subject independent manner. The first two parts were used for training, while the last part was used as a test set. The beginning and the end of each video correspond to neutral expressions. Accordingly, we used only the first half of the videos, which start from a neutral expression and ends with a peak expression.

\noindent \textbf{Extended Cohn Kanade (CK+)}~\cite{lucey2010extended}: This dataset comprises 327 sequences of posed expressions, annotated with seven expression labels from which we selected the six basic expressions. 
Each sequence starts with a neutral expression, and reaches the peak in the last frame. We used this dataset for data augmentation in the training of MotionGAN.  

\begin{figure*}
\centering
\includegraphics[width=17.33cm]{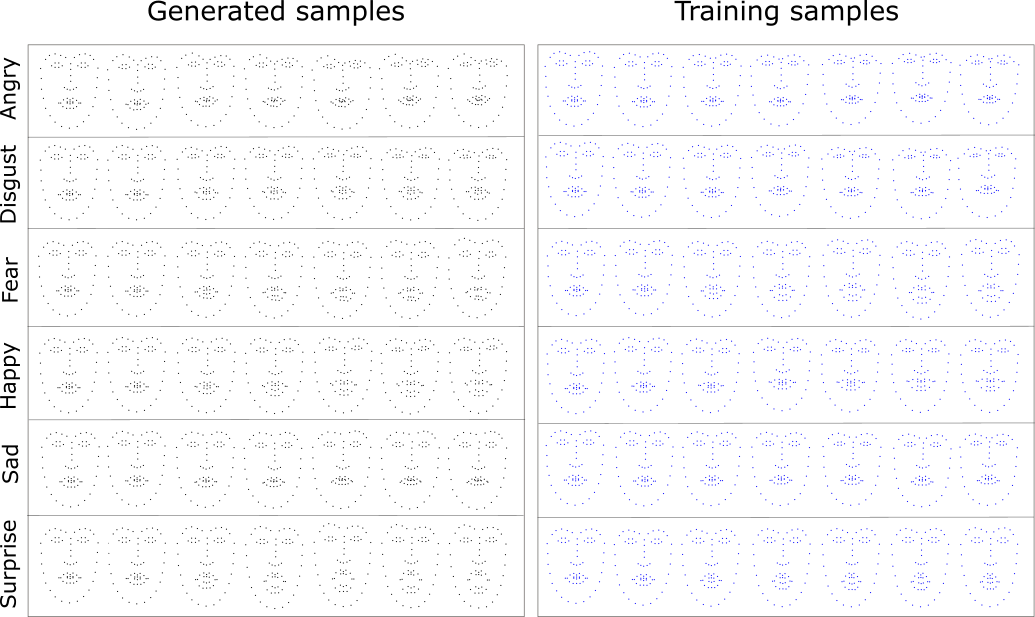}
\caption{Visualization of some facial landmark sequences. The left block shows landmark sequences obtained with the generated SRVF using MotionGAN applied to neutral landmark configurations. The right block shows some landmark sequences from the Oulu-CASIA dataset used in the training of MotionGAN. Each row corresponds to one facial expression. The original sequences contain 32 frames from which 7 frames were selected to visualize each sequence.}
\label{ShapeGAN_Qualitive_Results}
\end{figure*}

\subsection{Implementation Details}
\textbf{Preprocessing}: For all the images used in our experiments, we cropped the face regions using OpenFace~\cite{baltrusaitis2018openface}, and scaled them to $64 \times 64$. Then, we normalized all videos to $T=32$ frames using the approach proposed in~\cite{zhou2011towards}.
For MotionGAN data, we used OpenFace~\cite{baltrusaitis2018openface} to extract 2D coordinates of $68$ facial landmarks from each video frame. These landmarks were then arranged in a matrix representing a curve in $\mathbb{R}^{2d\times T}$, with $T=32$ representing the video length, and $d=68$ corresponding to the number of landmarks. We used~\eqref{curve_2_q} to compute the SRVF of the resulting curves. By computing the Karcher mean of the SRVFs belonging to the same class, we obtained a representative element of each expression class. Then, we aligned each training SRVF with the representative element of its corresponding class. The resulting SRVFs were used to train MotionGAN that can produce new SRVFs corresponding to new expression dynamics. The generated SRVFs were then transformed to landmark sequences using~\eqref{curve_2_q} with any neutral landmarks configuration. The training of TextureGAN was performed with pre-processed faces and guided by their ground truth landmark sequences, while the landmark sequences generated by MotionGAN were used in the test stage. 
Indeed, we avoid using landmark sequences generated by MotionGAN during TextureGAN training, since MotionGAN generates random motions for the same expression given that it starts from noise. Thus, we can not be sure to generate the exact sequence corresponding to the ground truth video that will be used to minimize the reconstruction loss of TextureGAN. To define the target pose for TextureGAN, we encoded landmark locations in heatmaps that were used as additional image channels to be concatenated with the input face image. Each heatmap is a multi-channel image with the same size as the input face image, where each channel encodes the location of one of the 68 landmarks, and the value of each pixel in a channel corresponds to the likelihood for its corresponding point location.

\textbf{MotionGAN architecture}: MotionGAN consists of multiple upsampling and downsampling blocks. The generator takes as input a vector of size 128 sampled from normal distribution and concatenated with the input label, which is encoded as one-hot vector of size 6 (number of classes). This vector is handled by one fully connected layer of $69,632$ outputs, and five upsampling blocks with 530, 274, 146, 64 and 1 output channels. Each upsampling block consists of the nearest-neighbor upsampling followed by a $3 \times 3$ stride $1$ convolution. The outputs of the first four convolution (Conv) layers are activated by the Relu function and concatenated with the label vectors that was transformed to six channels, while the last Conv layer uses hyperbolic tangent. The final output of the MotionGAN generator is a matrix of size $136 \time 32$ for each noise sample. 
The Discriminator of MotionGAN consists of three downsampling blocks with 64, 32 and 16 output channels. Each block is a $3 \times 3$ stride $1$ Conv layer followed by batch normalization and Relu activation. These layers are then followed by two fully connected (FC) layers of 1024 and 1 outputs. The first FC layer uses Leaky ReLU and batch normalization. Except the last FC layer, all outputs are concatenated with the label either in the form of six channels for Conv layers output or one-hot vector for the fully connected layers.

\textbf{TextureGAN architecture}: The TextureGAN generator takes in input a pre-processed face image concatenated with a heatmap encoding the target pose. We base our TextureGAN on the image-to-image translation network proposed in~\cite{pix2pix2017}. The generator is composed of an encoder and a decoder that have symmetric architectures. The encoder consists of six $4 \time 4$, $2$ stride Conv layers with 64, 128, 256, 1024, 1024 and 1024 output channels. Each Conv layer is followed by ReLu activation and batch normalization. Following the U-Network~\cite{ronneberger2015u} and~\cite{pix2pix2017}, we add skip connections between each layer $i$ and layer $l - i$, where $l$ is the number of the network layers. The skip connections consist of concatenating all channels at layer $i$ with those at layer $l - i$. To avoid outfighting, we add dropout with $0.5$ probability after the first three Conv layers. The architecture of the decoder is symmetric to that of the encoder. 

The discriminator of TextureGAN contains four $4 \times 4$ stride $2$ Conv layers with 64, 128, 256 and 1024 output channels. Each Conv layer is followed by batch normalization and Leaky ReLu activation except the first one that does not use batch normalization. The network ends with one fully connected layer of 1 output followed by sigmoid activation.

We trained the networks using the Adam optimizer~\cite{kingma2014adam}, with learning rate of 0.0002, and mini-batch size of 64 for TextureGAN and 128 for MotionGAN. Regarding MotionGAN, we empirically set its hyper-parameters to $\alpha_1= 0.8$, $\alpha_2= 1$ and $\alpha_3=1$, while we fix the hyper-parameters $\zeta_1=1$, $\zeta_2=1$ and $\zeta_3=80$ for TextureGAN. For data augmentation, we performed random flipping of the input images. The two networks MotionGAN and TextureGAN were trained separately, MotionGAN was trained for 200 epochs, while TextureGAN was trained for 400 epochs. We implemented our models with the Tensorflow~\cite{abadi2016tensorflow} framework based on the implementation of~\cite{pix2pix2017}. 
\subsection{Evaluation}
We assessed the performance of the proposed approach by: (\emph{i}) evaluating quantitatively and qualitatively the facial expression dynamics generated by MotionGAN; (\emph{ii}) assessing the quality of the videos generated by TextureGAN.   

\subsubsection{Landmark Sequence Generation}
\noindent \textbf{Qualitative results}: To qualitatively evaluate the generated expression dynamics (\emph{i.e.}, SRVF), we applied them to a neutral landmark configuration following~\eqref{eq:q_2_curve}. This results in sequences of landmarks that follow the dynamics encoded in the SRVFs. In Figure~\ref{ShapeGAN_Qualitive_Results}, we show some generated landmarks for the six basic expressions. The visualized landmark sequences show that MotionGAN is able to generate realistic expression dynamics from noise that are comparable with the ground truth sequences. Moreover, Figure~\ref{ShapeGAN_Qualitive_Results} shows that the proposed MotionGAN generator is able to synthesize expression dynamics (and their associated landmark configurations) corresponding to different conditioning labels. 

\noindent  \textbf{Quantitative Results}: To assess quantitatively the expression dynamics generated by MotionGAN and their associated landmark sequences, we propose to exploit the geodesic distance $d_\mathbb{S}$ between the SRVFs in $\mathbb{S}$ given by~\eqref{eq:distance_sphere}. This distance allows us to quantify the similarity between the generated sequences and those of the ground truth; it also allows us to measure the similarities between sequences of same or different expressions. 

\begin{figure}[!ht]
\centering
\subfloat[SRVF data from the Oulu-CASIA and CK+ datasets used in MotionGAN training]{
\label{subfig:correct}
\includegraphics[width=0.22\textwidth]{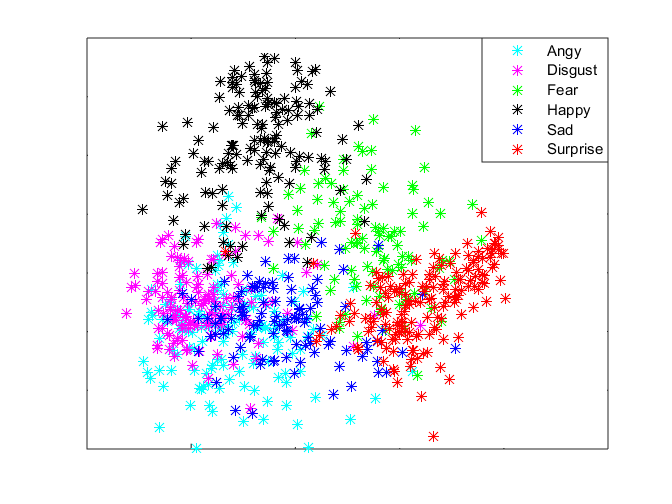}} \hspace{0.3cm}
\subfloat[SRVF data generated with MotionGAN]{
\label{subfig:notwhitelight}
\includegraphics[width=0.22\textwidth]{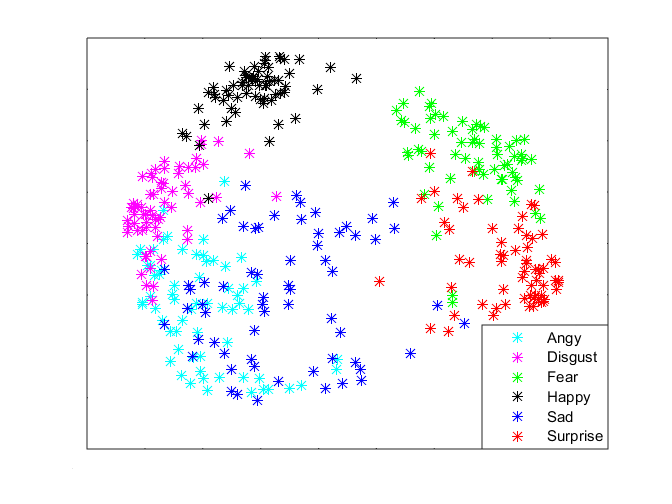}} 

\subfloat[Superposition of SRVF training and generated data]{
\label{subfig:nonkohler}
\includegraphics[width=0.22\textwidth]{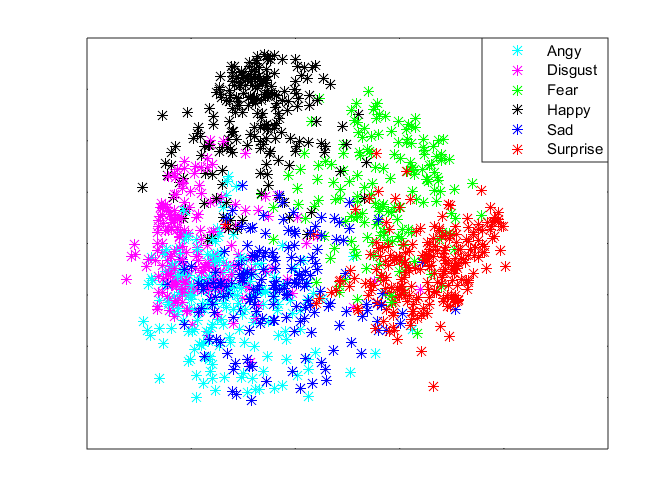}} 
\caption{2D visualization of the SRVF data using MDS based on the geodesic distance in $\mathbb{S}$.}
\label{fig:MSE_SRVF_visualization}
\end{figure}

Accordingly, we used the MotionGAN generator to synthesize 64 expression dynamics (\emph{i.e.}, SRVFs) for each one of the six basic expressions, which results in 384 generated \textit{SRVFs}. By computing the geodesic distances between all these generated samples, we used Multidimensional Scaling (MDS)~\cite{cox2000multidimensional} to visualize them in a 2D space. In Figure~\ref{fig:MSE_SRVF_visualization}, we show in 2D the generated samples as well as those of the databases used to train MotionGAN. The first aspect that can be noted is the effectiveness of the representation used to encode the motion of the facial expressions. Indeed, the plots show that SRVFs can easily differentiate expression classes, while keeping close the classes that share more inter-class similarities (\emph{e.g.}, Fear and Surprise). 
We also notice that most of the generated expressions are well separated, which demonstrates that MotionGAN learns the distribution of each class and it is capable of generating samples conditioned on the input labels. Moreover, this visualization shows the diversity of the generated motions, which allows us to generate different dynamics for the same expression.

\begin{figure*}[!ht]
\centering
\includegraphics[width=17.00cm]{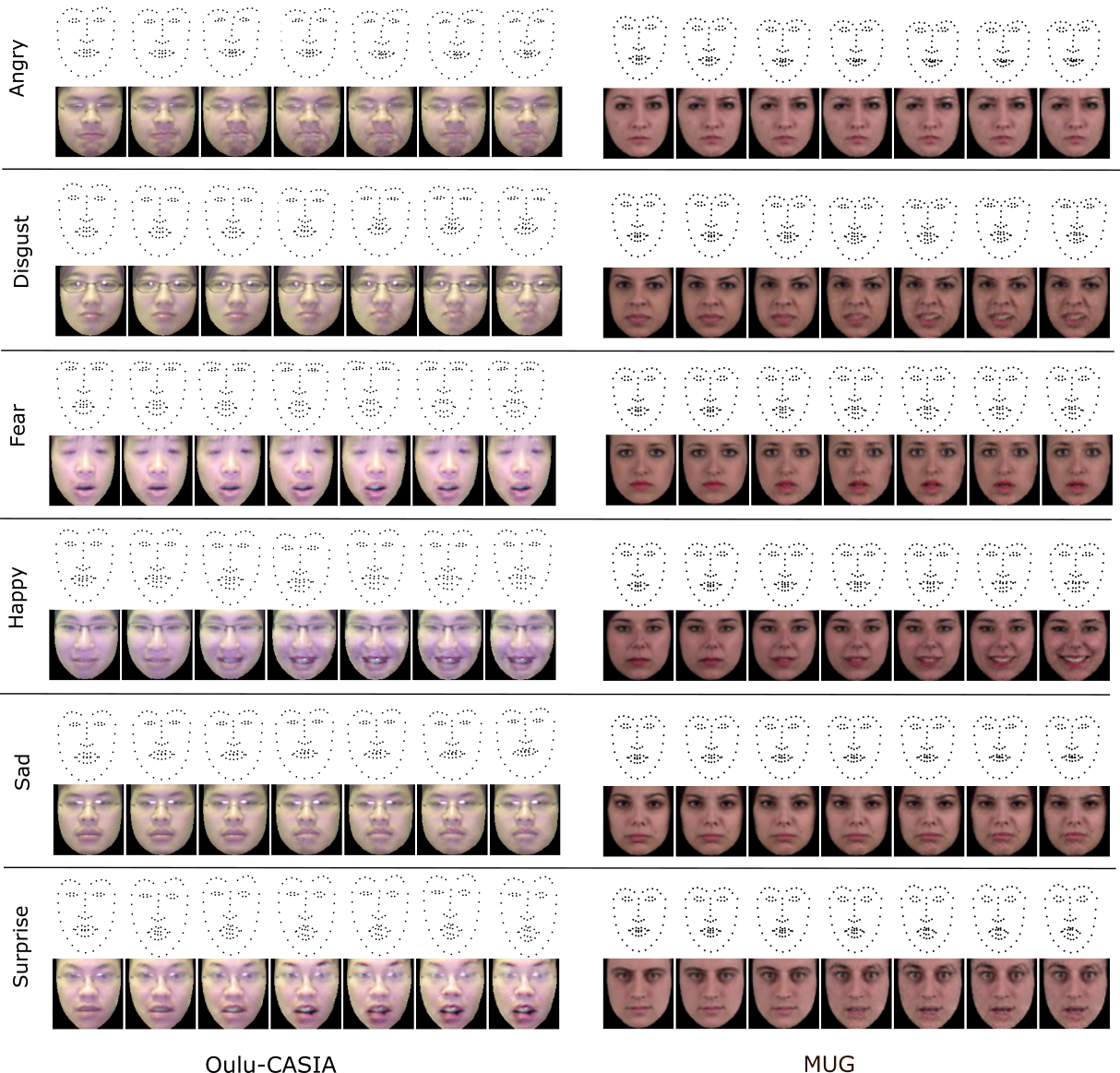}
\caption{Generated videos by MotionGAN and TextureGAN. The image sequences were randomly selected and the reported images are taken every 5 frames. Some examples of animated videos are shown in this \href{https://drive.google.com/drive/folders/1R2zHvtvWfDKKaNgCTXKiua-3-oqWmO5O?usp=sharing}{link}.}
\label{fig:pix2pixResults}
\end{figure*}

\subsubsection{Video Generation}
In this section, we aim at evaluating the quality of the final generated videos. These videos are guided by landmark sequences resulting from applying the generated SRVFs on neutral input landmark configurations.

\noindent  \textbf{Qualitative Results:} In Figure~\ref{fig:pix2pixResults}, we show some videos generated with our approach. Firstly, from noise, we generate expression dynamics using MotionGAN, then we apply these dynamics to the landmarks configuration of the face at hand. This results in a sequence of landmarks that is used to guide TextureGAN to generate the final video from the input face image. All the faces used in the following results are taken from the test set of the corresponding dataset. We also note that the split of the datasets was subject independent, thus all the used faces here are not seen by the model during the training phase. 
The resulting videos demonstrate that our method is able to synthesize realistic videos of previously unseen faces with smooth continuity and without discrete transitions. Moreover, the results show that our model preserves well the characteristics of the input faces such as identity and eyeglasses. It can also generate the teeth region, which does not exist in the source image for happy and surprise expressions. Some examples of animated videos are shown at this \href{https://drive.google.com/drive/folders/1R2zHvtvWfDKKaNgCTXKiua-3-oqWmO5O?usp=sharing}{link}.

\begin{figure*}
\centering
\subfloat[Generated by MoCoGAN~\cite{Tulyakov2018}]{\includegraphics[scale=.51]{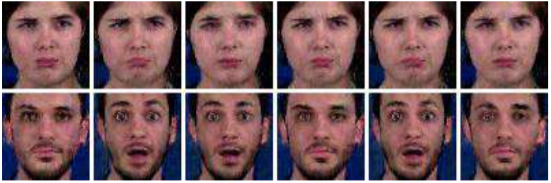}}
\hfill
\subfloat[Generated by VGAN~\cite{NIPS2016_6194}]{\includegraphics[scale=0.51]{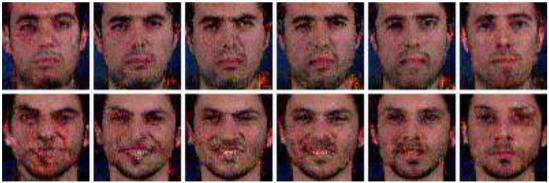}}
\hfill
\subfloat[Generated by TGAN~\cite{MasakiSaito2017}]{\includegraphics[scale=0.51]{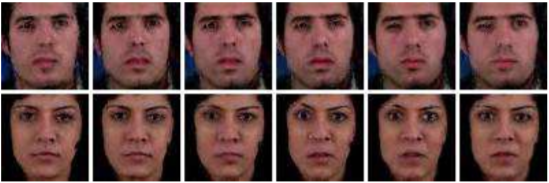}}
    
\subfloat[Generated by MotionGAN and TextureGAN]{\includegraphics[scale=0.565]{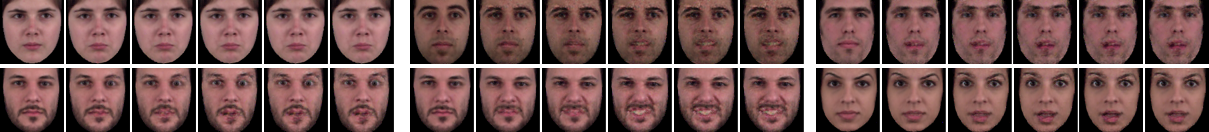}}
\vspace{0.4cm}
\caption{Qualitative comparison with the state-of-the-art on the MUG Facial Expression database. The samples generated by our model are randomly selected, while those generated with state-of-art approaches are taken from~\cite{Tulyakov2018}. 
}
\label{fig:Qualitative_Comparison}
\end{figure*}

In Figure~\ref{fig:Qualitative_Comparison}, we compare some frames generated by our model with three state-of-the-art video generation approaches: MoCoGAN~\cite{Tulyakov2018}, VGAN~\cite{NIPS2016_6194}, and TGAN~\cite{MasakiSaito2017}. 
For each approach, in the two first rows, we show the synthesized videos of two identities performing two expressions. Then, we report our generated videos for the same identities and the same expressions in the last two rows. We observe that our model can generate more realistic images with less content and motion artifacts, while other approaches generate blurry images and failed to keep fine details of the expressions. Even the dynamics of our generated videos looks more natural and changes more smoothly compared to that generated by the other approaches.

\NO{Indeed, the SRVF representation allows us to generate the motion of the video as one point on the hypersphere manifold, which minimizes motion artifacts and results in smoother changes over time. This is one advantage of our approach compared to methods based on RNN that generate each time step separately. This smoothness is well preserved when transforming the generated SRVF to landmark sequences. By using landmark sequences to guide the generation of a final video, we preserve smoothness, since each frame corresponds to its associated landmark configuration in the smooth landmarks sequence.}

\noindent \textbf{Quantitative Results}: \NO{To evaluate our model quantitatively, we used different metrics commonly used in the state-of-the-art: Inception Score, PSNR, SSIM, ACD and ACD-I. 
The Inception Score (IS) proposed in~\cite{gulrajani2017improved} is a common used metric to assess the quality of the generated samples that correlates very well with human judgment. The computation of this metric involves using a labeled dataset and a good expressions classifier. Accordingly, we used the MUG database and we trained a Long Short-Term Memory Network (LSTM)~\cite{hochreiter1997long} model with one layer for expression classification on this dataset. The input features of the LSTM network consist of expression features learned by the CNN model used in~\cite{otberdout2019automatic, Otberdout0DBB18}. This CNN was trained for expression recognition from static images on the training set of the Oulu-CASIA dataset. Using this CNN, we extracted vectors of $128$ features from the last fully connected layer for each video frame. Then, we used these feature vectors as inputs to the LSTM model to classify the whole video. 
The Peak Signal-to-Noise Ratio (PSNR) and Structural Similarity (SSIM)~\cite{wang2004image} are metrics that reflect the quality of the generated samples by measuring pixel-level similarity between the generated videos and their ground truth. 
The last metric that we used to evaluate the quality of the generated examples is the Average Content Distance (ACD)~\cite{Tulyakov2018} and its ACD-I variant proposed in~\cite{zhao2018learning}. ACD measures the content consistency of the generated video based on how well the video preserves identity of the input face. We computed this metric by using OpenFace~\cite{amos2016openface}, which is a deep model trained for face recognition. This model extracts identity features in the form of vectors from each frame of the video. Then, the ACD is given by the average of the distances between vectors of consecutive frames. The ACD-I metric is the average distance between each generated frame and the original input frame. It is used to assess how the facial identity is captured in the generated video. 
All the reported results in Table~\ref{SOTA_comparison} were obtained by generating videos of six basic facial expressions for each subject of the test sample of the MUG facial expressions dataset, which consists of a total number of 1400 videos. We further report the results on the training set as reference for IS, ACD and ACD-I metrics. 

Regarding the compared methods VGAN and MoCoGAN, we used the public code provided by the authors with minor changes as explained below. 
In order to compare our results with those of VGAN~\cite{NIPS2016_6194}, we trained the conditional version of VGAN that uses an encoder to encode the input neutral image. Then, the resulting latent code of the input image is concatenated with the desired label encoded as a one-hot vector, and feed to the generator of VGAN. While our method and VGAN 
generate videos from neutral frame of a given identity, MoCoGAN samples videos from a random content.   All the compared models here were trained from scratch on the MUG dataset including MotionGAN and TextureGAN. 

The quantitative results reported in Table~\ref{SOTA_comparison} show that our approach outperforms other existing methods. The IS achieved by our approach is the best over the compared methods, which proves the high diversity and quality of our generated samples. Furthermore, our method attains the best PSNR and SSIM scores, which is consistent with the IS results. Regarding the ACD and ACD-I scores, we achieve the best values demonstrating that our approach is capable of generating videos that well preserve the identity of the input face, which is also consistent with the qualitative results discussed previously. 

To further quantitatively assess the quality of our generated samples, we reported in Table~\ref{Recognition_Rate} the recognition rate of an expression classifier on randomly generated samples. To this end, we exploited the LSTM model described above for IS computation. The table includes the recognition rate of this model on: the test set of the MUG database as an upper bound, as well as randomly generated samples with our model, VGAN and MoCoGAN. Note that the test set of the MUG dataset contains $106$ samples, thus in order to provide a fair comparison, all the reported results were obtained with $106$ generated samples using the identities of the MUG test set.
While the recognition rate on the real samples surpasses the classification rate achieved with our generated samples by $6.59\%$, our approach attains the best recognition rate comparing to VGAN and MoCoGAN with a difference of $24.54\%$ and $16.82\%$, respectively. These results are consistent with the qualitative comparison provided in Figure~\ref{fig:Qualitative_Comparison} showing that our generated samples are more realistic and present less artifacts than VGAN and MoCoGAN. This helps the CNN model used for features extraction to identify the expression patterns in each frame of the video and allows the LSTM model to recognize the motion of the expression. Results of this experiment prove that our method generates realistic samples both in term of images and motions, that can be recognized with a dynamic expressions classifier.}

\begin{table}[!t]
\caption{\NO{Comparison with state-of-the-art for using Inception Score (IS), PSNR, Structural Similarity (SSIM), Average Content Distance (ACD) and ACD-I. The line named Reference accounts for real sequences in the MUG test set. PSNR and SSIM are not reported for MoCoGAN since it samples from a random content}}
\label{SOTA_comparison}
\centering
\NO{
\begin{tabular}{|l|c|c|c|c|c|}
\hline
\textbf{Approach} & \textbf{IS} & \textbf{PSNR} & \textbf{SSIM} & \textbf{ACD} & \textbf{ACD-I} \\
\hline\hline
VGAN~\cite{NIPS2016_6194} & $3.24\pm 0.59$ & $20.98$ & $0.80$ & $0.051$ & $0.162$ \\
\hline
MoCoGAN~\cite{Tulyakov2018} & $3.10\pm 0.38$& -&-& $0.045$& $0.14$ \\
\hline
{\bf Our} & \textbf{$3.52 \pm 0.55$} & \textbf{$25.90$} & \textbf{$0.90$} & \textbf{$0.011$} &  \textbf{$0.120$}\\
\hline \hline
Reference & $4.48 \pm 0.64$ & Inf & $1.00$ &$0.009$ & $0.084$  \\
\hline
\end{tabular}}
\end{table}

\begin{table}[!t]
\renewcommand{\arraystretch}{1.}
\caption{\NO{Comparison with state-of-the-art in expression recognition using an LSTM on the generated samples. The line named Reference accounts for real sequences in the MUG test set}}
\label{Recognition_Rate}
\centering
\NO{
\begin{tabular}{|l|c|}
\hline
\textbf{Approach} &  \textbf{Accuracy (\%)} \\
\hline\hline
VGAN~\cite{NIPS2016_6194} & $45.28$ \\
\hline
MoCoGAN~\cite{Tulyakov2018} & 53.00\\
\hline 
{\bf Our} & $69.82$ \\
\hline\hline
Reference & $76.41$ \\
\hline
\end{tabular}}
\end{table}

\begin{figure}
\centering
\includegraphics[width=7.00cm]{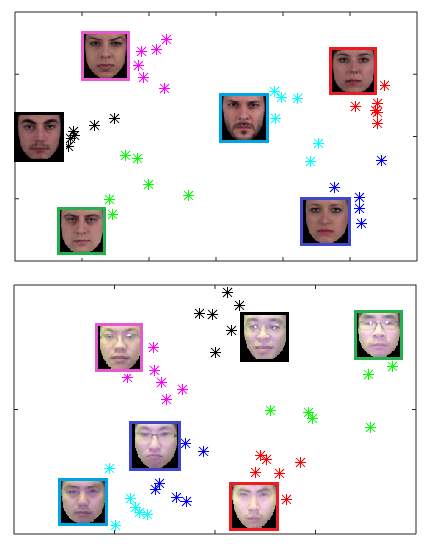}
\caption{2D visualization of the identity features space of six subjects chosen randomely from the MUG (top) and Oulu-CASIA (bottom) datasets. The neutral face images in the plot show the identity of the subjects, while the colored dots indicate the generated expressive ones.}
\label{fig:Identity_features}
\end{figure}

\noindent
\textbf{Identity features visualization}: Here, we visualize the identity features of the generated videos. To this end, we used the OpenFace~\cite{amos2016openface} model to generate identity features for each frame of the videos. The average Euclidean distance between these features is then used with MDS to visualize, in a 2D space, the identity features of the generated videos of six identities chosen randomly. Figure~\ref{fig:Identity_features} supports our previous results and shows that our model preserves the identity information of the input face.


\subsection{Ablation Study}
In order to evaluate the effect of each component of our model, we conducted an ablation study on the Oulu-CASIA dataset. In this study, we exploited the four metrics introduced above: PSNR, SSIM, ACD and ACD-I. 
The ablations study was performed using three models. Each model was trained by avoiding one component of the full model. In the first model, we discarded the identity loss during the training of TextureGAN, while, in the second model, we avoided the reconstruction loss on $\mathbb{S}$ during the training of MotionGAN. The last model consists of using ground truth landmark sequences to guide TextureGAN instead of those generated by \textit{MotionGAN}. 
Results are shown in Table~\ref{fig:ablation_study}. From these results, we notice that using the identity loss during TextureGAN training improves the content consistency of the generated videos as indicated by the higher ACD and ACD-I scores. This is expected as the ACD and ACD-I metrics are based on the similarities of identity features. This result demonstrates that using identity loss helps to maintain the subject identity throughout the generated videos. The model trained without reconstruction loss in $\mathbb{S}$ shows the worst PSNR and SSIM scores and higher ACD and ACD-I comparing to our full model, which demonstrates the usefulness of this loss in training a good landmark sequences generator that leads to good videos quality. 
Regarding the third model that uses ground truth landmarks to guide TextureGAN, we notice that it achieves the best scores, while the results are still close to those achieved by our full model. This evidences that MotionGAN is capable of generating landmark sequences that are very similar to those of the ground truth. 

\begin{table}[!t]
\renewcommand{\arraystretch}{1.}
\caption{Ablation study on the Oulu-CASIA dataset}
\label{fig:ablation_study}
\centering
\begin{tabular}{|l|c|c|c|c|}
\hline
\textbf{Model} & \textbf{ACD} & \textbf{ACD-I} & \textbf{PSNR} & \textbf{SSIM} \\
\hline\hline
w/o $L_{id}$ & 0.0193 & 0.143 & \textbf{26.05} & \textbf{0.908} \\
\hline
w/o $L_{S\_recon}$ & 0.017 & 0.127 & 24.180 & 0.886 \\
\hline
w/o \textit{MotionGAN} & \textbf{0.016} & \textbf{0.10} & 25.98 & 0.90 \\
\hline
Full model & \textbf{0.016} & 0.107 & 24.443 & 0.891 \\
\hline
\end{tabular}
\end{table}

\subsection{Data Augmentation for Expression Recognition}
In order to quantitatively demonstrate the usefulness of the synthesized expression videos, we trained an expression classifier based on an LSTM with one layer. This LSTM model was used to classify the videos of the test set of the CASIA dataset. The input features of the LSTM network consist of expression features learned by the CNN model used in the IS computation and described previously. 
We first trained the LSTM on the training set, and considered the accuracy achieved by this model on the test set as baseline. Then, we re-trained the model from scratch by augmenting the training data. In each experiment, we multiplied the number of the original training data by 4, 6, 10 and 15. The generated data used for data augmentation here correspond to the same identities of the original training data performing new motions generated by our MotionGAN. 
\NO{In order to assess the diversity of the generated motions by MotionGAN, we further report the results of the same experiment without MotionGAN. This is achieved by transferring motions of the real videos to different identities instead of using motions generated by MotionGAN.} 

\NO{Results obtained in these experiments are reported in Table~\ref{tab:data_augmentation}. We notice that multiplying the number of training data by 4, 6 and 10 using our generated samples improves the results from $87.5\%$ to $90.62\%$, $91.66\%$ and $92.7\%$, respectively. Instead, results saturate when multiplying the number of the original data by more than 10 times. These results show the usefulness of our approach in generating new samples comparable with real ones that can be exploited for video data augmentation to train improved emotion recognition models.
Comparing these results with those obtained with real motions, we notice that the results are mostly similar, which proves the high quality of the generated motions. We can explain the similar results between using MotionGAN and omitting it by the fact that the model is saturated in $92.7\%$ and the motions of the test set are very close to the motions of the training set. Therefore, even if MotionGAN can generate higher diversity than that of the dataset, we still get similar results for the test set. Furthermore, we can notice that using MotionGAN, the model saturated with less data (\emph{i.e.}, by multiplying the data by $10$) comparing to the model that omits MotionGAN, which necessitates to multiply the data by $15$ to saturate.}
 

\begin{table}[!ht]
\renewcommand{\arraystretch}{1.}
\caption{\NO{Expression recognition (\%) obtained by training an LSTM on the CASIA original training set (baseline), and the original training set augmented with an increasing number of synthesized videos. Both the cases where MotionGAN or the Real Motion is used for generating the landmark dynamics have been tested}}
\label{tab:data_augmentation}
\centering
\NO{
\begin{tabular}{|l|c|c|}
\hline
\textbf{\# Training samples} & \textbf{MotionGAN } & \textbf{Real Motions}\\
\hline\hline
CASIA training data (baseline)  &  87.5 & 87.5\\
\hline\hline
\# Training data $\times 4$ & 90.62 &  91.66\\
\hline
\# Training data $\times 6$ &  91.66 & 91.66\\
\hline
\# Training data $\times 10$ & 92.7 & 91.66\\
\hline
\# Training data $\times 15$ & 92.7 & 92.7 \\
\hline
\end{tabular}}
\end{table}

\subsection{Facial Expression Transfer}
\begin{figure}[!ht]
\centering
\includegraphics[width=9.1cm]{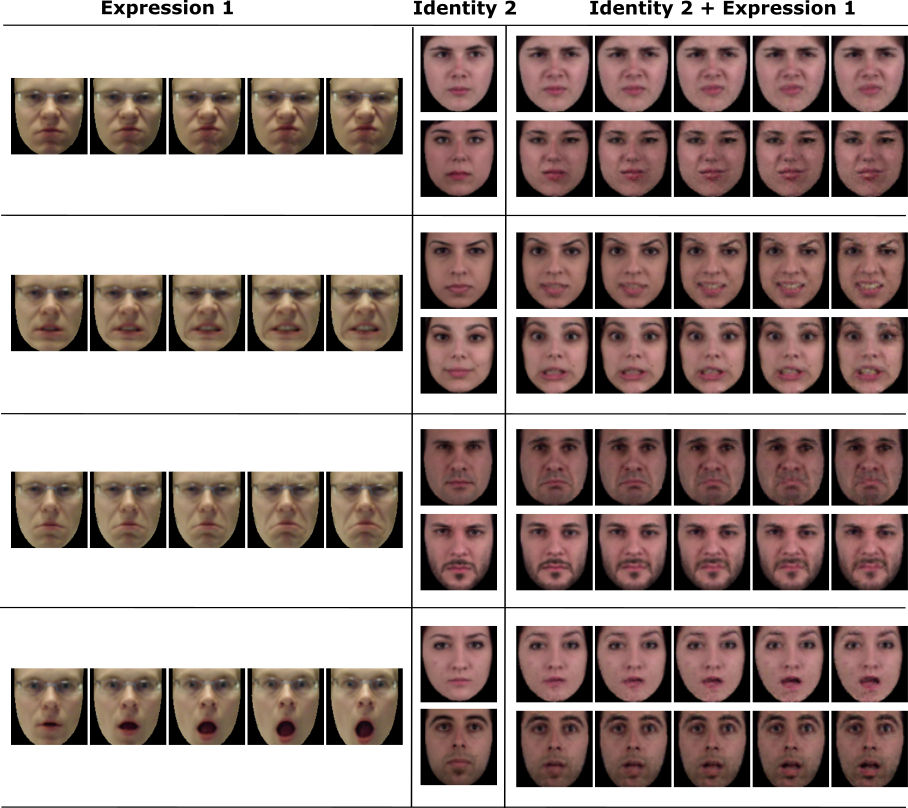}
\caption{Facial expression transfer. Expressions from the left column taken from the Oulu-CASIA dataset are transferred to faces in the middle column taken from the MUG dataset. Each expression was transferred to two different identities and the results of the transfer are shown in the right column.}
\label{fig:transfert}
\end{figure}

Facial expression transfer aims to transfer expressions from a source subject to a target one. The newly-synthesized expressions of the target subject are supposed to be identity-preserving and exhibit similar emotions to the source subject. 
In addition to facial expressions synthesis, our proposed model can also be used for dynamic facial expression transfer. In Figure~\ref{fig:transfert}, we show that our model is able to transfer a dynamic expression $Expr_1$ from a source identity $Id_1$ to a target one $Id_2$. This is achieved by encoding the motion of $Id_1$ in a \textit{SRVF} representation, then using~\eqref{eq:q_2_curve} we map this motion to the neutral landmarks configuration of the target identity $Id_2$. Finally, the resulting landmark sequences are used to guide \textit{TextureGAN} to generate the final video of $Id_2$ performing the dynamic expression $Exp_1$ of $Id_1$. In Figure~\ref{fig:transfert}, we used an identity from the Oulu-CASIA database performing four facial expressions; disgust, fear, sad and surprise. Then, we transfer each one of these expressions to two identities in the MUG dataset. Results show that our model is able to transfer the facial expressions to different identities with good image quality, while keeping the characteristics of the target identity. For what concerns facial expression generation, we can notice that also in expression transfer the model can synthesize the teeth region that was hidden in the input face. 

\section{Conclusion}\label{sect:conclusions}
In this paper, we addressed the difficult task of dynamic facial expression generation given a neutral face image. We proposed a novel framework, which processes separately facial expression dynamics and face appearance using two different GAN architectures. The temporal information of the video was first represented by the facial landmarks evolution that was encoded as a point on the Hypersphere manifold. We proposed a conditional version of manifold-valued Wasserstein GAN that has been used to generate new facial expression motions corresponding to a given emotion state. Finally, the second conditional real-valued GAN transforms the generated facial landmark sequences to video frames by adding the texture information. We evaluated the proposed approach quantitatively by using different metrics commonly used in the state-of-the-art: Inception Score, PSNR, SSIM, ACD and ACD-I. We showed that the proposed approach significantly outperforms state-of-the-art in video face generation. The reported experiments on two public datasets demonstrate the effectiveness of our approach in facial expression editing, facial expression transfer and data augmentation. In our future works, we aim to generalize our approach to deal with 3D facial expressions and 3D actions generation.

\section*{Acknowledgements}
This work was supported by CNRST's scholarship of Excellence (Morocco), and by CAMPUS FRANCE [PHC TOUBKAL 2019 (French-Morocco Bilateral Program)] under Grant 41539RH. It was partially supported by the French State, managed by the National Agency for Research (ANR) under the Investments for the future program with reference ANR-16-IDEX-0004 ULNE. 


%





\ifCLASSOPTIONcaptionsoff
  \newpage
\fi




%



\ifCLASSOPTIONcaptionsoff
  \newpage
\fi

\vskip -2\baselineskip plus -1fil

\begin{IEEEbiography}[{\includegraphics[width=1in,height=1.25in,clip,keepaspectratio]{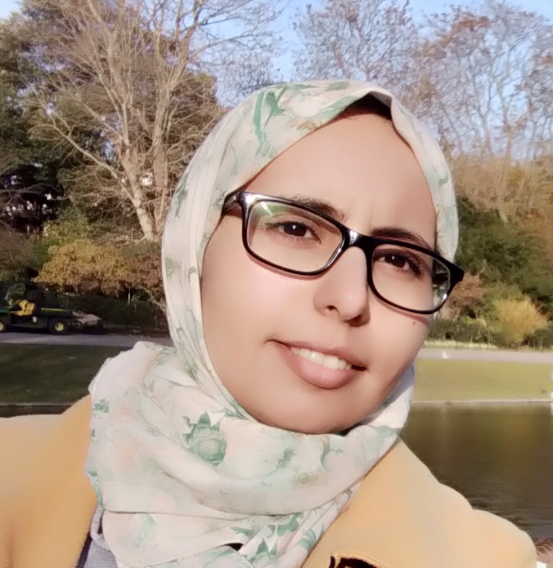}}]{Naima Otberdout} received her master's degree in computer sciences and telecommunication in 2016, from Mohammed V University in Morocco. Currently, she is a Ph.D. candidate in Computer Science at the same University. Her general research interest is computer vision and machine learning. Particularly, she is interested in deep learning with Riemannian geometry for face analysis and human behavior understanding.
\end{IEEEbiography}

\vskip -2\baselineskip plus -1fil

\begin{IEEEbiography}[{\includegraphics[width=1in,height=1.25in,clip,keepaspectratio]{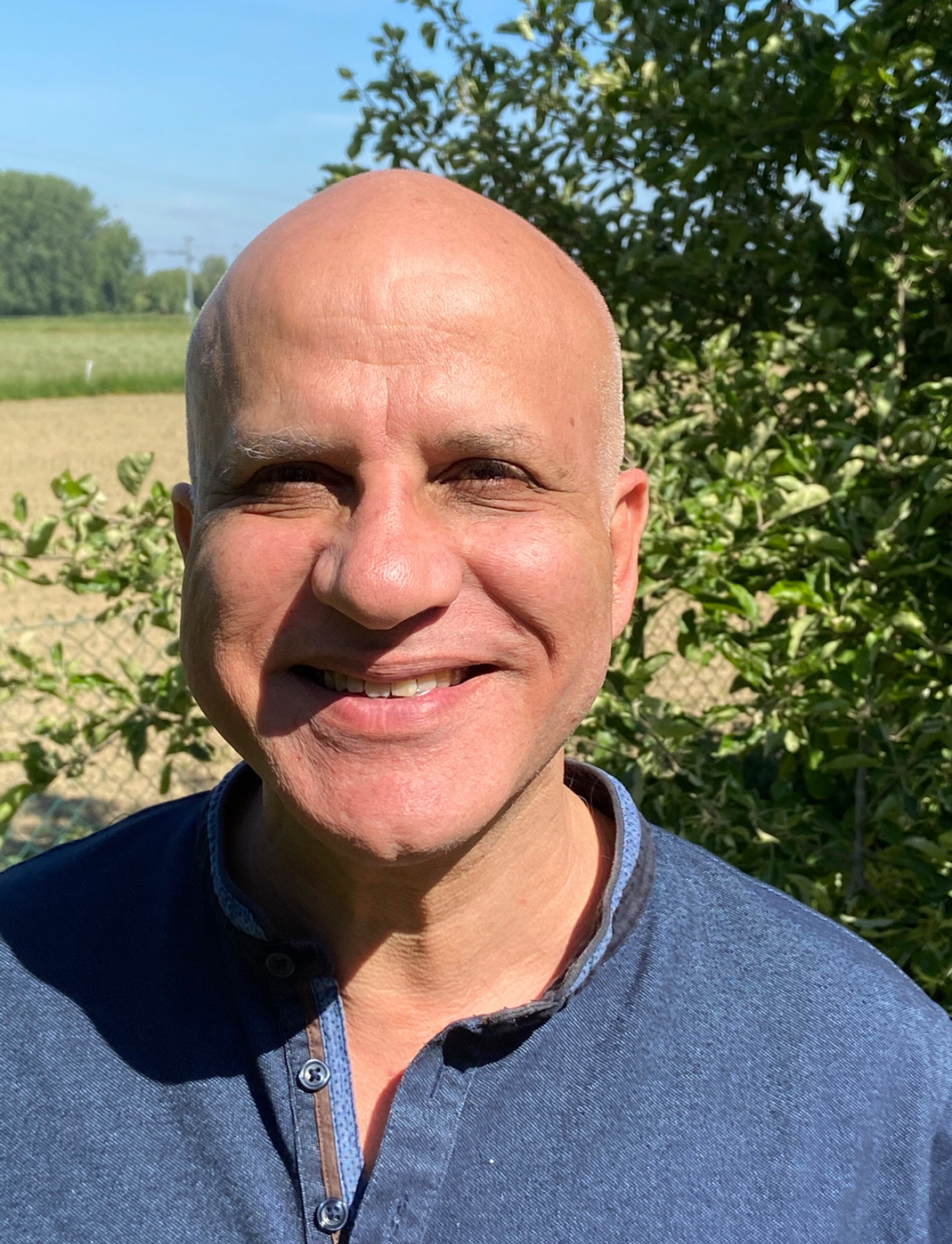}}]{Mohamed Daoudi} is a Full Professor of Computer Science at IMT Lille Douai and the Head of Image group at CRIStAL Laboratory (UMR CNRS 9189). He received his Ph.D. degree in Computer Engineering from the University of Lille (France) in 1993 and Habilitation A Diriger des Recherches from the University of Littoral (France) in 2000. His research interests include pattern recognition, shape analysis and computer vision. He has published over 150 papers in some of the most distinguished scientific journals and international conferences. He is Associate Editor of Image and Vision Computing Journal, IEEE Transactions on Multimedia and Journal of Imaging. He has served as General Co-Chair of IEEE International Conference on Automatic Face and Gesture Recognition, 2019. He is Fellow of IAPR and IEEE Senior member.
\end{IEEEbiography}
 
\vskip -2\baselineskip plus -1fil

\begin{IEEEbiography}[{\includegraphics[width=1in,height=1.25in,clip,keepaspectratio]{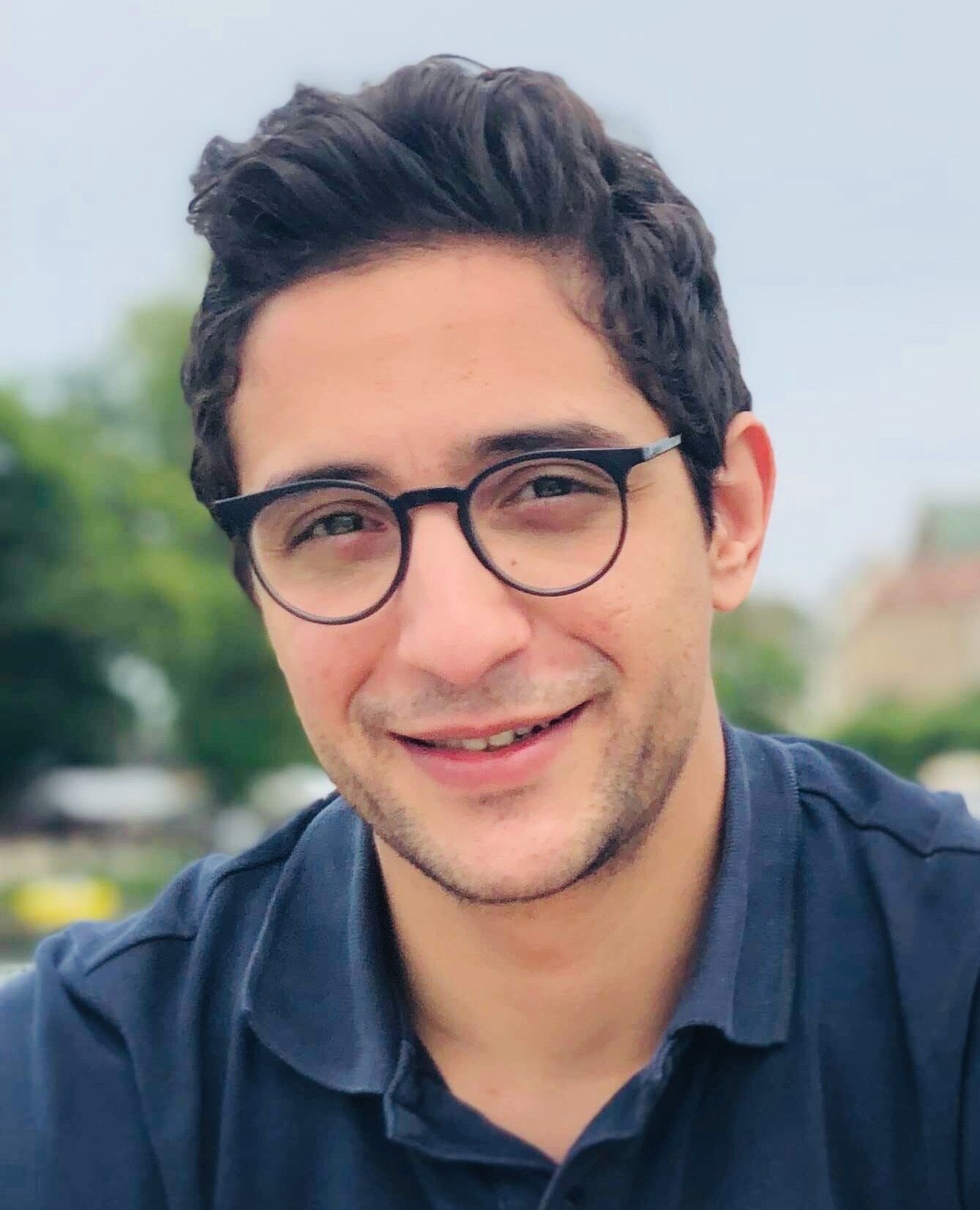}}]{Anis Kacem} received his Ph.D. degree in computer science from the Institut Mines Teleecom (IMT) Lille-Douai and University of Lille in France, within the 3DSAM research team of CRIStAL laboratory (CNRS UMR 9189). Currently, he is a research associate at Interdisciplinary Centre for Security, Reliability and Trust (SnT) in University of Luxembourg. His research interests are mainly focused on computer vision and pattern recognition with applications to human behavior understanding.
\end{IEEEbiography}
 
\vskip -2\baselineskip plus -1fil
 
\begin{IEEEbiography}[{\includegraphics[width=1in,height=1.25in,clip,keepaspectratio]{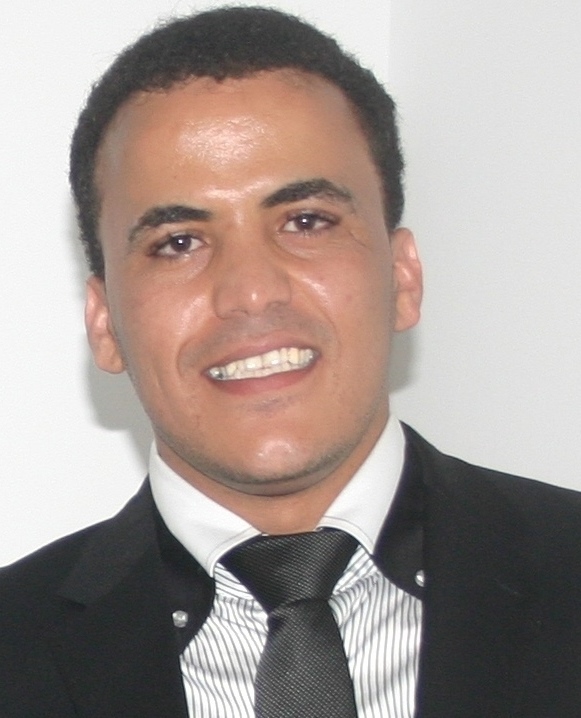}}]{Lahoucine Ballihi} is an Associate Professor at the Department of Computer Science in Faculty of Science, Mohammed V University in Rabat, Morocco. He received his Master of Advanced Studies (MAS) degree in Computer science and telecommunications from Mohammed V University in Rabat, Morocco in 2007 and the Ph.D degree in Computer Science from the University of Lille, France and the Mohammed V University in Rabat, Morocco in 2012.
He is a member of the Computer Science and Telecommunications Research Laboratory of Mohammed V University (LRIT - CNRST URAC 29). His current research interests include computer vision, machine learning, biometrics, three-dimensional analysis and retrieval, image and video analysis and categorization, face and action analysis and recognition, and affective computing. He has published over 25 papers in some of the most distinguished scientific journals and international conferences
\end{IEEEbiography}

\vskip -2\baselineskip plus -1fil

\begin{IEEEbiography}[{\includegraphics[width=1in,height=1.25in,clip,keepaspectratio]{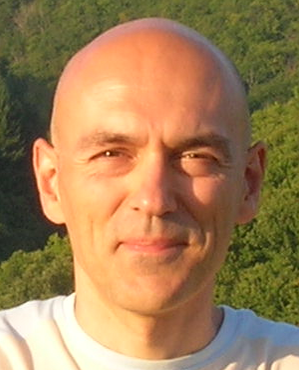}}]{Stefano Berretti} received the Ph.D. in Computer Engineering in 2001. Currently, he is an Associate Professor at University of Florence, Italy. He has been Visiting Professor at the University of Lille and the University of Alberta.
His research interests focus on computer vision for face biometrics, human emotion and behavior understanding, computer graphics and multimedia. 
He is the Information Director and an Associate Editor of the ACM Transactions on Multimedia Computing, Communications, and Applications, and Associate Editor of the IET Computer Vision journal.
\end{IEEEbiography}








\section*{Supplementary Material to the Paper: Dynamic Facial Expression Generation on Hilbert Hypersphere with Conditional Wasserstein Generative Adversarial Nets}

%
%
\author{Naima~Otberdout,~\IEEEmembership{Member,~IEEE,}
Mohamed~Daoudi,~\IEEEmembership{Senior,~IEEE,}
Anis~Kacem,~\IEEEmembership{Member,~IEEE,}
Lahoucine~Ballihi,~\IEEEmembership{Member,~IEEE,}
and~Stefano~Berretti,~\IEEEmembership{Senior,~IEEE}
\IEEEcompsocitemizethanks{\IEEEcompsocthanksitem N. Otberdout and L. Ballihi are with the LRIT - CNRST URAC 29, Mohammed V University in Rabat, Faculty of Sciences, Rabat, Morocco,
E-mail: naima.otberdout@um5s.net.ma, lahoucine.ballihi@um5.ac.ma 
\IEEEcompsocthanksitem A. Kacem and M. Daoudi are with IMT Lille-Douai, University of Lille, CNRS, UMR 9189 CRIStAL, Lille, France. E-mail:. \{anis.kacem,mohamed.daoudi\}@imt-lille-douai.fr
\IEEEcompsocthanksitem S. Berretti is with the Department of Information Engineering, University of Florence, 
Florence, Italy. E-mail: stefano.berretti@unifi.it}}
\maketitle



%

In this supplementary material, we provide more visual results of our approach to show the diversity of the motions generated by MotionGAN. 
In addition, we show that using our approach (\emph{i})  it is possible to control the intensity of the generated facial expressions, (\emph{ii}) the generated motions can be also applied to the landmarks of non-frontal faces, and (\emph{iii}) a qualitative comparison between sequences generated by our method and MoCoGAN.

\subsection*{Visualization of the identity space of the generated samples with VGG features}
\begin{figure}[!ht]
\centering 
\includegraphics[width=\linewidth]{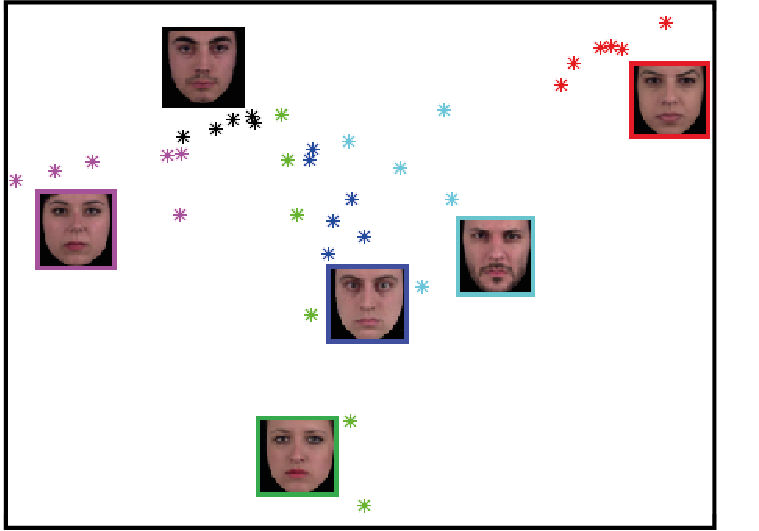}
\caption{2D visualization of the identity VGG feature space of six subjects chosen randomly from the MUG dataset. The neutral face images in the plot show the identity of the subjects, while the colored dots indicate the generated expressive ones.}
\label{Fig:VGG_features}
\end{figure}

In our proposed approach, VGG features were used to preserve identity during TextureGAN training. So, in Figure~\ref{Fig:VGG_features} we visualize the VGG identity features of the generated videos. 
To this end, we used the VGG-face~\cite{parkhi2015deep} model to generate identity features for each frame of the videos. The average Euclidean distance between these features is then used with multidimensional scaling to visualize, in a 2D space, the identity features of the generated videos of the same six identities used in Figure~8 in the main paper. This figure supports our previous results and shows that our model preserves the identity information of the input face.

\subsection*{Diversity of the generated samples by MotionGAN}
\begin{figure*}
\begin{center}
\includegraphics[width=19.0cm]{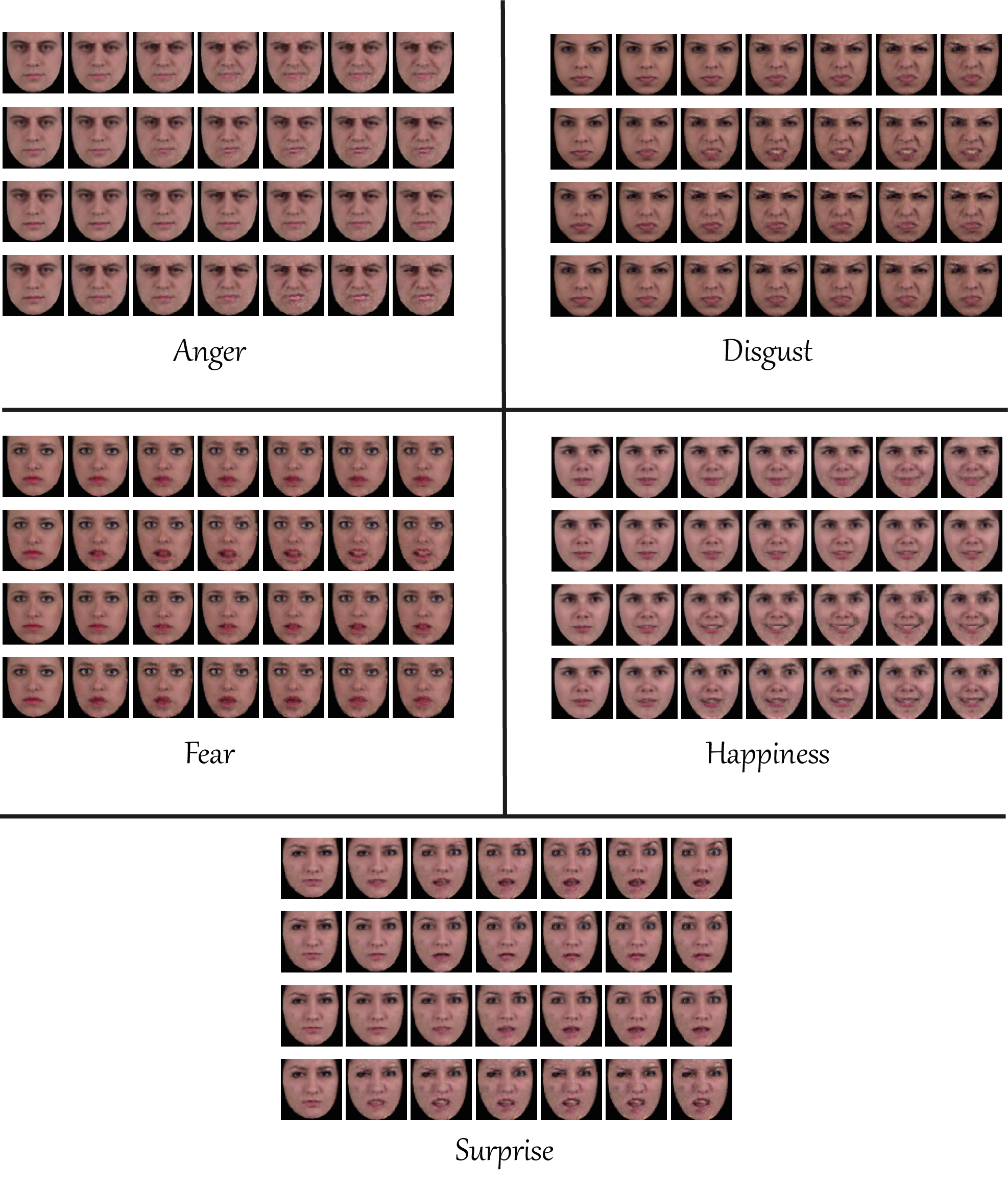}
\end{center}
\caption{Four different sequences of each facial expression performed by the same identity. Each box corresponds to one facial expression performed differently by the same identity. In these boxes, each row shows one video generated by our model (\emph{i.e.}, MotionGAN and TextureGAN). The reported images of each video are taken every 5 frames. It is clearly visible that MotionGAN can generate diverse motions for the same facial expression, while keeping the common average pattern of the expression between them.}
\label{Fig:Diversity}
\end{figure*}

In the main paper, we demonstrated the diversity of the generated motions in Figure 5-b, where the 2D visualization of some generated samples based on the geodesic distances between them is reported. This figure shows that the model is capable of distinguishing different class labels, while generating different samples distant from each other in the same class label. To further show the diversity of the generated motions, we visualize in Figure~\ref{Fig:Diversity} some identities performing various generated motions by MotionGAN belonging to the same facial expression class. This figure shows that MotionGAN can generate different motions for the same facial expression, while keeping the common patterns related to this expression. The complete videos associated to this figure can be found at the following link\footnote[1]{https://sites.google.com/view/tpamisi-2019-07-0608/accueil}.

In addition to the diversity of the generated motions, our approach can easily control the intensity of the generated facial expressions. Given that the motion generated by MotionGAN can be regarded as controlled changes applied to the initial facial landmarks as expressed in~(3) in the main paper, we can change the intensity of the facial expression by multiplying these changes by a factor $I$ such that:
\begin{equation}
\beta(t) = I \cdot \int_{0}^{T}\|q(s)\| q(s) ds + \beta(0) \; ,
\label{eq:q_2_curve}
\end{equation}

\begin{figure*}
\begin{center}
\includegraphics[width=19.0cm]{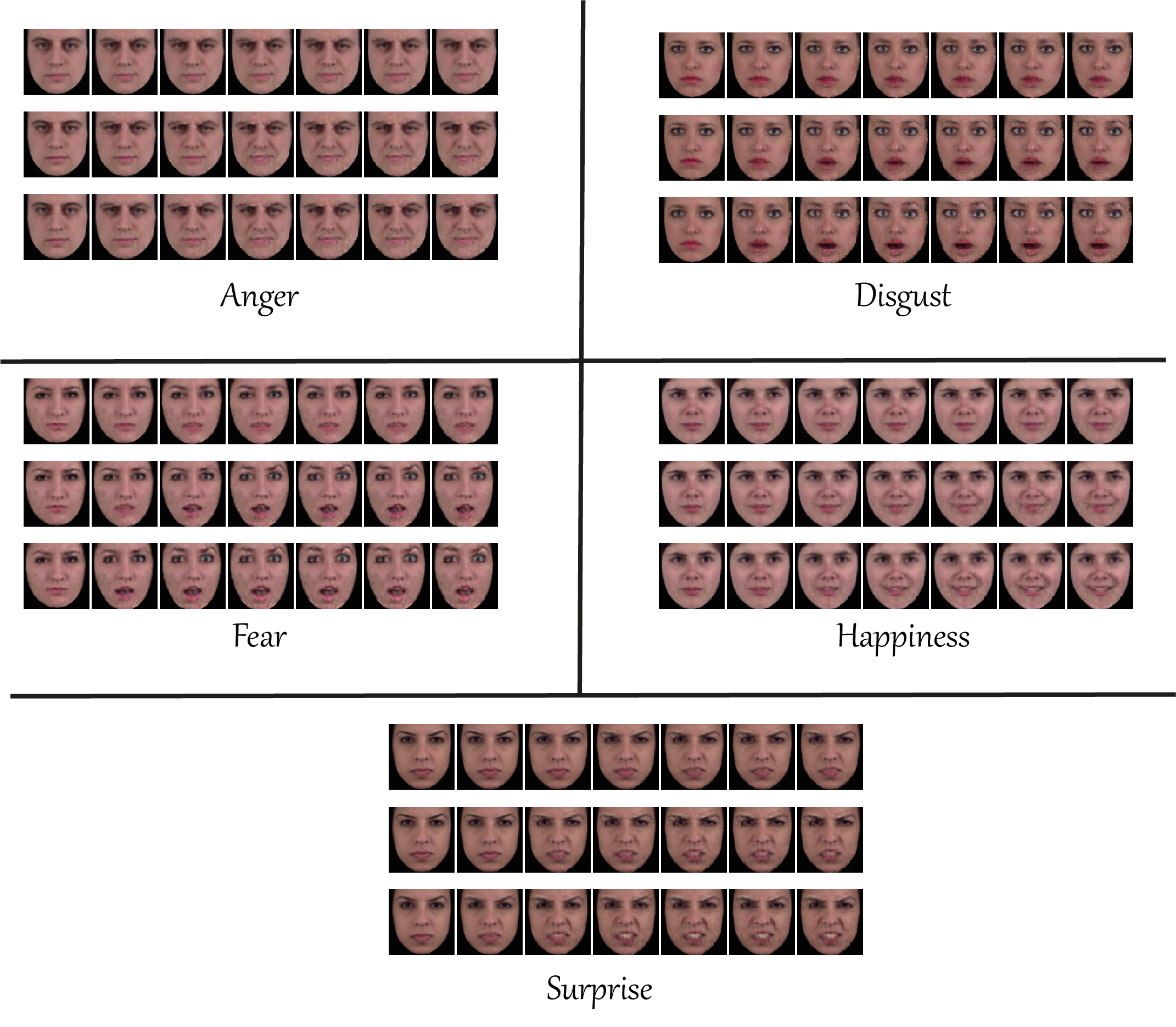}
\end{center}
\caption{Different intensities of the same motion generated by MotionGAN for the same identity. We reported one box for each facial expression. In these boxes, each row shows the video generated by our model for the same motion but different intensities (value of $I$ in Eq.(3)) equal to 10, 20, and 30, respectively). 
The images reported for each video are taken every 5 frames. The complete videos associated to this figure can be found at the following \href{https://sites.google.com/view/tpamisi-2019-07-0608/accueil}{link}. }
\label{Fig:Intensity}
\end{figure*}

\noindent 
By controlling this factor, we were able to generate the same facial expression corresponding to the same generated motion with different intensities. Results are shown in Figure~\ref{Fig:Intensity}. In these visualisations, we set the factor $I$ to $10$, $20$ and $30$; Using greater values for $I$ we obtained implausible expressions. 

\subsection*{MotionGAN for non-frontal faces}
\begin{figure*}
\begin{center}
\includegraphics[width=18.0cm]{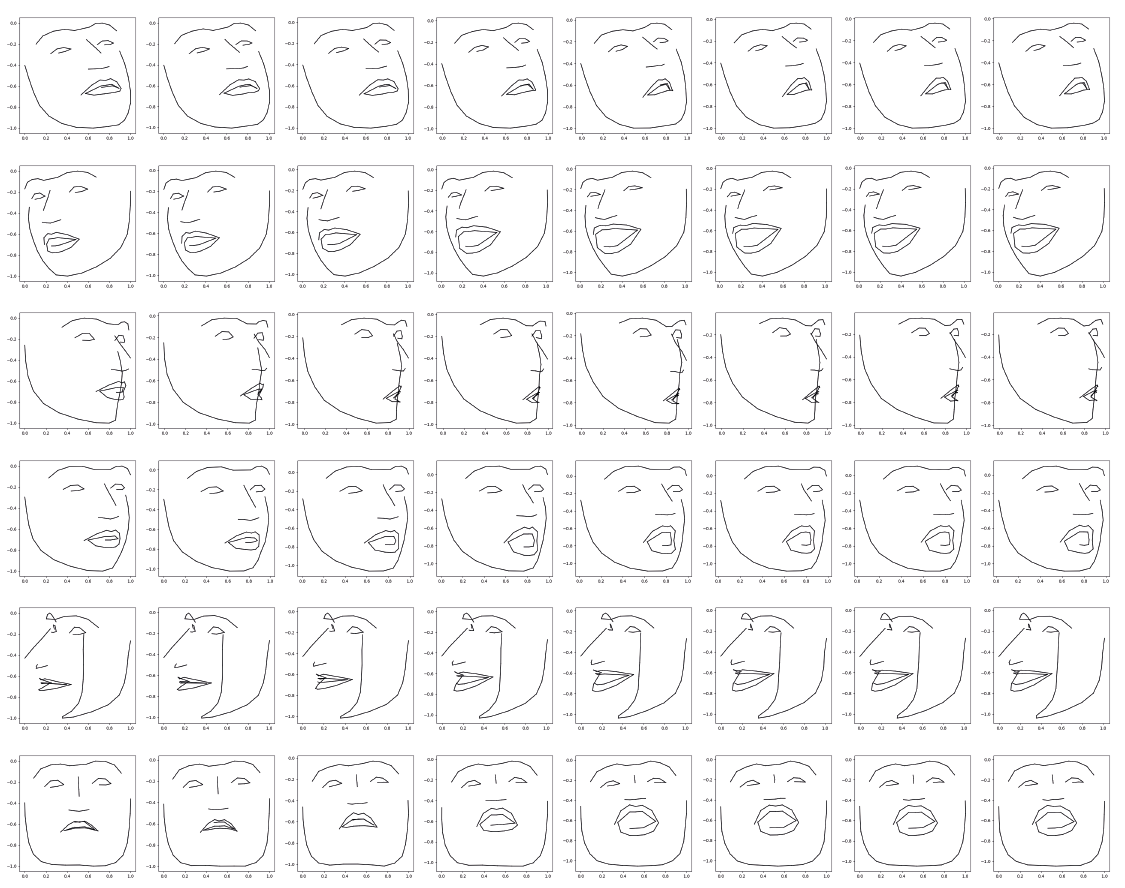}
\end{center}
\caption{Visualization of some facial expression motions generated by MotionGAN applied to non-frontal faces. Each row shows seven frames of one video. The first left column corresponds to neutral faces.}
\label{Fig:non_frontal_faces}
\end{figure*}

In this paper, we chose to deal with 2D frontal facial expression generation. However, the motions generated by MotionGAN can be easily applied to non-frontal facial landmark configuration. In Figure~\ref{Fig:non_frontal_faces}, we show some motions generated by MotionGAN applied to non-frontal faces. 

\subsection*{Qualitative Comparison with MoCoGAN}
\begin{figure*}
\begin{center}
\includegraphics[width=18.0cm]{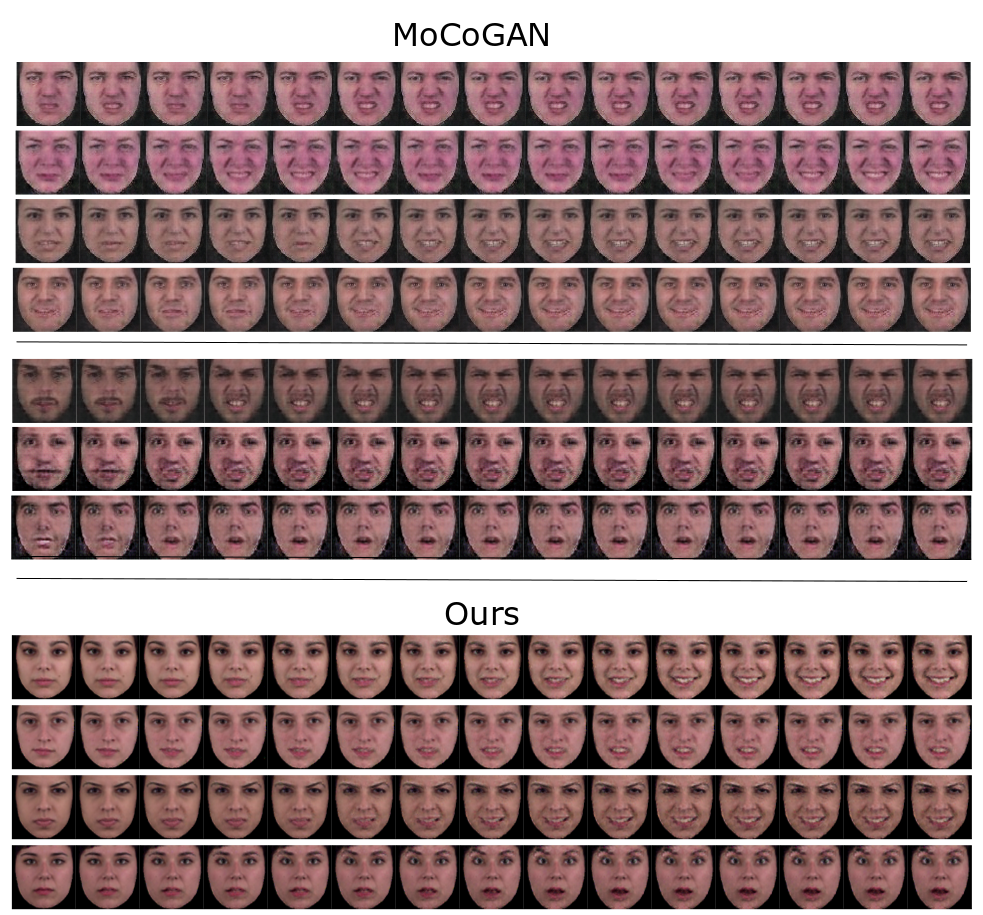}
\end{center}
\caption{Qualitative comparison between our generated samples and those of MoCoGAN. We notice that MoCoGAN generates many videos presenting temporal artifacts. In the first group of MoCoGAN samples, the intensity of the expression does not increase continuously. The person starts the facial expression than return to the neutral state before continuing the expression. The second group of MoCoGAN shows sudden changes, where the videos start with neutral face that switch suddenly to the peak expression. In our generated samples, we did not 
observe these artifacts. Our generated samples present smoother and continuous changes. Note that we did not visualize the same identities for the two methods since MoCoGAN generates samples from a random content, so we cannot control the identity of the generated videos.}
\label{Fig:artifactss}
\end{figure*}

We visualize in Figure~\ref{Fig:artifactss} some generated samples using our model (MotionGAN and TextureGAN) along with other examples generated by MoCoGAN. This figure shows that our method generates more realistic examples and eliminates some temporal artifacts presented in the MoCoGAN samples. 

\end{document}